\documentclass[10pt]{article}

\usepackage[margin=1in]{geometry}

\usepackage[utf8]{inputenc}
\usepackage[T1]{fontenc}
\usepackage{amsmath,amssymb,amsthm}
\usepackage{hyperref}
\usepackage{url}
\usepackage{booktabs}
\usepackage{graphicx}
\usepackage{float}
\usepackage{natbib}
\usepackage{enumitem}
\usepackage{subcaption}

\usepackage{amssymb} 
\usepackage[most]{tcolorbox}
\tcbuselibrary{listings}
\usepackage[table]{xcolor}
\usepackage[most]{tcolorbox}
\tcbuselibrary{listings}
\usepackage{tikz}
\usetikzlibrary{positioning}
\usetikzlibrary{arrows.meta,positioning,fit}
\usetikzlibrary{decorations.pathreplacing}
\usetikzlibrary{calc}

\newtcolorbox{promptbox}{
    enhanced,
    colback=gray!4!white,
    colframe=gray!60!black,
    boxrule=0.5pt,
    arc=4pt,
    left=8pt, right=8pt, top=10pt, bottom=8pt,
    title=\textbf{Model Prompt},
    coltitle=black,
    attach boxed title to top left={xshift=12pt, yshift=-10pt},
    boxed title style={colback=gray!15!white, colframe=gray!60!black, boxrule=0.5pt, arc=2pt},
    breakable,
    before=\par\vspace{1em}\noindent,
    after=\par\vspace{1em}
}

\newtcolorbox{compactprompt}[1]{%
    enhanced,
    colback=gray!4!white,
    colframe=gray!60!black,
    boxrule=0.5pt,
    arc=3pt,
    left=7pt, right=7pt, top=4pt, bottom=4pt,
    title={\textbf{#1}},
    coltitle=black,
    fonttitle=\small,
    attach boxed title to top left={xshift=8pt,yshift=-2pt},
    boxed title style={
        colback=gray!15!white,
        colframe=gray!60!black,
        boxrule=0.4pt,
        arc=2pt,
        left=5pt,
        right=5pt,
        top=1pt,
        bottom=1pt
    },
    fontupper=\small,
    parbox=false,
    before=\par\vspace{0.5em}\noindent,
    after=\par\vspace{0.3em}
}

\newtcblisting{codebox}{
    enhanced,
    colback=blue!3!white,
    colframe=blue!40!black,
    boxrule=0.5pt,
    arc=3pt,
    left=8pt, right=8pt, top=6pt, bottom=6pt,
    listing only,
    listing options={
        language=Python,
        basicstyle=\ttfamily\small,
        keywordstyle=\color{blue!80!black}\bfseries,
        commentstyle=\color{gray!80!black}\itshape,
        stringstyle=\color{green!50!black},
        showstringspaces=false,
        breaklines=true,
        tabsize=4
    },
    breakable,
    before=\par\vspace{0.5em}\noindent,
    after=\par\vspace{0.5em}
}

\newcommand{\ding}[1]{\ifnum#1=51 $\checkmark$\else$\times$\fi}

\usepackage{xspace}
\newcommand{\horizonmath}{\textbf{HorizonMath}\xspace}

\title{Measuring AI Progress Toward Mathematical Discovery with Automatic Verification}

\usepackage{authblk}

\author{%
\normalfont\normalsize
Erik Y.~Wang$^{1,2,*}$,
Sumeet R.~Motwani$^{3,*}$,
James V.~Roggeveen$^{4}$,
Eliot~Hodges$^{5}$,
Dulhan~Jayalath$^{3}$,
Charles~London$^{3}$,
Kalyan~Ramakrishnan$^{3}$,
Jakob~Foerster$^{3}$,
Cheng~Zhang$^{6}$,
Flaviu~Cipcigan$^{6}$,
Philip~Torr$^{3}$,
Alessandro~Abate$^{3}$\\
{\normalfont\small
$^{1}$Stanford University \quad
$^{2}$Benchmark \quad
$^{3}$University of Oxford\\
$^{4}$Harvard University \quad
$^{5}$Princeton University \quad
$^{6}$Ellison Institute of Technology
}%
}

\begin{document}

\date{}
\maketitle
\begingroup
\makeatletter
\renewcommand{\@makefnmark}{}
\footnotetext{%
$^{*}$ Indicates joint-first authors. Correspondence to \href{mailto:erikwang@stanford.edu}{\texttt{erikwang@stanford.edu}} and \href{mailto:sumeet.motwani@eng.ox.ac.uk}{\texttt{sumeet.motwani@eng.ox.ac.uk}}. The benchmark and its instructions are available at the project page: \url{https://ewang26.github.io/HorizonMath/}.}
\makeatother
\endgroup
\vspace{-1em}

\begin{abstract}

  Can AI make progress on important, unsolved mathematical problems? Large language models are now capable of sophisticated mathematical and scientific reasoning, but whether they can perform novel research is still widely debated and underexplored. We introduce \horizonmath, a benchmark of 113 predominantly unsolved problems spanning eight domains in mathematics and the mathematical sciences, paired with an open-source evaluation framework for automated verification. Our benchmark targets the generator-verifier gap: problems where discovery is hard and requires meaningful mathematical insight, but verification is computationally straightforward. This contrasts with most existing research-level benchmarks, which instead rely on formal proof verification or manual review, both of which are expensive to scale. Because these solutions are unknown, HorizonMath is resistant to data contamination, and most state-of-the-art models score under 10\%. Using this framework, we identify six novel solutions to research problems that either resolve previously open questions or improve on the best-known published results, with GPT-5.4 Pro and GPT-5.6 Sol each discovering three of these solutions. Across seven frontier model families, reasoning efficiency and behavior also vary substantially. We release HorizonMath as an open challenge and a growing community resource, where each verified solution is a candidate contribution to the mathematical literature.

\end{abstract}

\section{Introduction}
\label{sec:intro}
Making autonomous mathematical discoveries is a north star in AI research, as mathematics is a key driver of scientific discovery. FunSearch~\cite{romera2024mathematical} and AlphaEvolve~\cite{georgiev2025mathematical} improved several open bounds in combinatorics, geometry, and number theory; frontier LLM-based agents have resolved previously unsolved Erdős problems~\cite{feng2026semi, sothanaphan2026resolution}; and human-AI collaboration has produced new results across theoretical computer science, physics, and pure mathematics~\cite{knuth2026claude, guevara2026single, woodruff2026accelerating}. As these capabilities continue to advance, standardized methods are needed to systematically measure AI progress toward autonomous mathematical discovery.

Many mathematics benchmarks exist, but most were designed to test \emph{problem-solving} rather than \emph{discovery}, which requires assessing novel mathematical contributions. 
This problem-solving framing of applying known techniques to problems with known answers has a natural ceiling—human knowledge—and even the most challenging Olympiad-style and graduate-level benchmarks~\cite{luong2025towards, tsoukalas2024putnambench, roggeveenhardmath2, rein2024gpqa} evaluate problems with known solutions. As such, there exist very few benchmarks that can provide signal on whether AI systems can perform research and produce results not already in the literature. 

\begin{figure}
    \centering
    \includegraphics[width=0.90\linewidth]{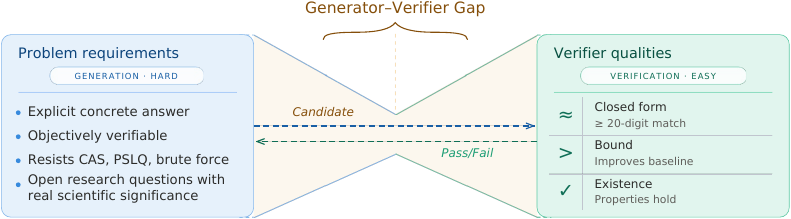}
    \caption{The \emph{generator--verifier gap} of \horizonmath. Problems require hard mathematical discovery, while submitted candidates are checked by deterministic verifiers. Formal proofs are not required for any problems in the dataset.}
\label{fig:principles}
\end{figure}

On the other hand, measuring capabilities on unsolved problems is \emph{naturally difficult} because solutions are unknown, but it is crucial since it requires genuinely novel mathematical contributions. We therefore exploit the generator-verifier gap, a feature in certain classes of problems in which candidate solutions are hard to produce but efficient to check. We identify three such forms that are well-suited to automated evaluation: 1) problems where a closed-form solution is currently unknown, but a candidate expression can be checked against a high-precision numerical reference, 2) construction and optimization problems seeking objects that improve upon a current baseline not known to be optimal, and 3) existence problems where a target object has not been found but a candidate can be validated by checking that it satisfies all required properties. All three classes can be efficiently verified by deterministic computation, making them natural targets for a scalable benchmark of mathematical discovery (Figure~\ref{fig:principles}).

\begin{figure}
    \centering
    \includegraphics[width=0.98\linewidth]{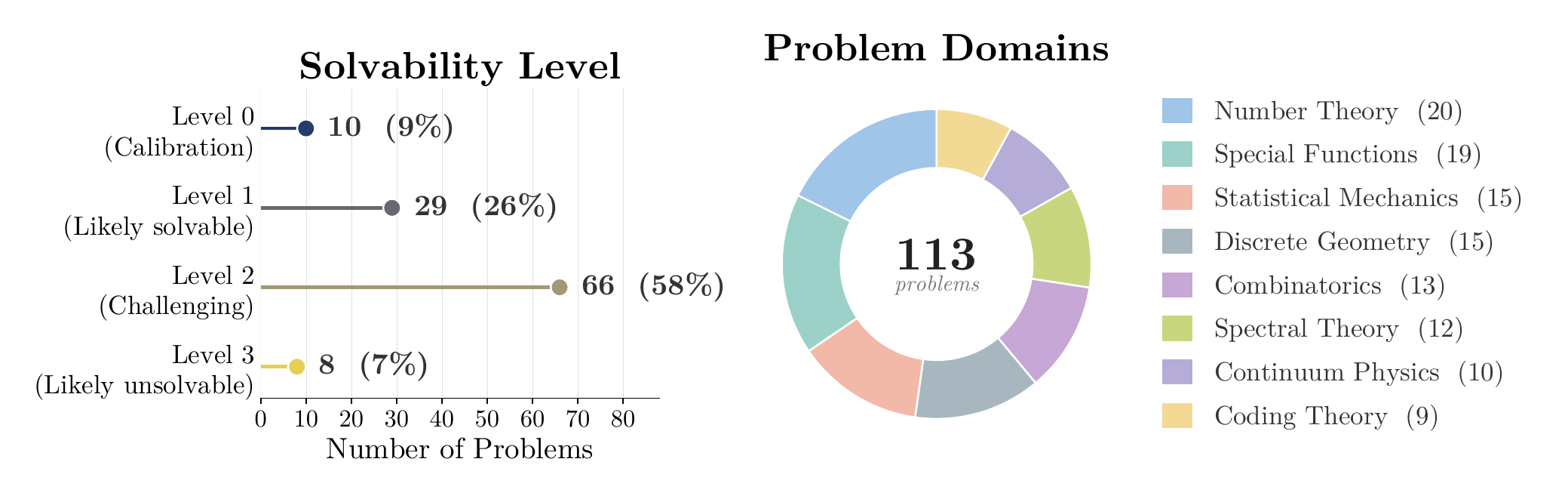}
    \caption{Benchmark composition by solvability level (left) and mathematical domain (right). HorizonMath comprises 113 problems, distributed across several areas of mathematics and physics.}
    \label{fig:benchmark-composition}
\end{figure}

With these classes of problems, we present \horizonmath, a benchmark of 113 predominantly unsolved problems and associated verification scripts that automatically evaluate proposed answers against problem-specific criteria (Figure~\ref{fig:benchmark-composition}). Our framework is defined by three key contributions:

\begin{enumerate}[leftmargin=*]
\item \textbf{A Contamination-Resistant Benchmark of Open Problems:} We present a benchmark of 113 problems across eight subject areas in applied and computational mathematics. Because the solutions are unknown, they do not exist in any training corpus, and any correct solution produced by a model would therefore signal genuine reasoning ability and autonomous discovery.

\item \textbf{Automated Verification:} Evaluations of mathematical research capabilities are traditionally bottlenecked by human review. We automate verification using high-precision numeric comparison and deterministic constraint-checkers, using the generator-verifier gap to provide a fast and objective signal supporting correctness.

\item \textbf{Open-Source and Standardized Evaluation:} Unlike some other benchmarks of research-level mathematics, our framework includes complete problem definitions, ground-truth computations, and verifier scripts, and is publicly available.
\end{enumerate}

We note that while matching a high-precision numerical reference does not formally prove that a closed-form expression is exactly correct, our combination of numerical comparison and verification of permitted operations provides significant evidence for correct solutions and can effectively filter inadmissible ones. Using GPT-5.4 Pro and GPT-5.6 Sol Pro, we identify three novel solutions from each model that either resolve previously open questions or improve on the best-known published results; these six results are described in Section \ref{sec:Results} and detailed in Appendix \ref{sec:Appendix}. \horizonmath is an open-source effort that welcomes contributions from the community and is available at: \url{https://github.com/ewang26/HorizonMath}.

\section{Related work}
\label{sec:related}




We organize related work into three categories that highlight the contrast between solution and discovery, introduced in Section~\ref{sec:intro}. We first review \emph{solution-driven} mathematical benchmarks, which evaluate models on problems with known answers (Section~\ref{sec:related-solution}). We then survey \emph{research-level} benchmarks, which target unsolved problems and help define our design space, and also note the key differences between these works and \horizonmath (Section~\ref{sec:related-research}). Finally, we discuss recent demonstrations of AI-driven mathematical discovery, which establish the capabilities that \horizonmath seeks to measure (Section~\ref{sec:related-discovery}).

\subsection{Solution-driven mathematical benchmarks}
\label{sec:related-solution}

Early benchmarks like GSM8K~\cite{cobbe2021training} and MATH~\cite{hendrycks2measuring} are now saturated, and while recent benchmarks targeting higher difficulty levels, such as GPQA~\cite{rein2024gpqa} and the HARDMath series~\cite{fanhardmath, roggeveenhardmath2} for graduate-level problems have emerged, these benchmarks all evaluate problems with \emph{known} solutions and therefore cannot measure novel discovery. Benchmarks based on competitions and Olympiads (e.g., OlympiadBench~\cite{he2024olympiadbench}, IMO-Bench~\cite{luong2025towards}, and MathArena~\cite{balunovicmatharena}) have provided new and difficult problems but are typically conducted on an annual basis and do not readily support continuous measurement of research progress, especially as they are all already-solved problems.

\definecolor{light-blue}{rgb}{0.88, 0.95, 1.0}
\newcommand{\greencheck}{{\color{green!60!black}\ding{51}}}
\newcommand{\redcross}{{\color{red}\ding{55}}}

\begin{table}[H]
\centering
\setlength{\tabcolsep}{4pt}
\caption{Comparison of math reasoning benchmarks.}
\label{tab:comparison}
\resizebox{\textwidth}{!}{%
\begin{tabular}{lcccccc}
\toprule
\textbf{Dataset} & \textbf{Problems} & \textbf{Difficulty} & \textbf{Unsolved} & \textbf{Non-Proof-Centric} & \textbf{Open-evaluation} & \textbf{Auto-verify} \\
\midrule
FrontierMath \cite{glazer2024frontiermath} & 300+ & Research & \redcross & \greencheck & \redcross & \greencheck \\
FrontierMath: Open Problems \cite{epoch2026frontiermath_open} & 14 & Research & \greencheck & \greencheck & \redcross & \greencheck \\
IMProofBench \cite{schmitt2025improofbench} & 39 & Research & \redcross & \redcross & \redcross & \redcross \\
First Proof \cite{abouzaid2026first} & 10 & Research & \redcross & \redcross & \redcross & \redcross \\
Erd\H{o}s Problems \cite{Bloom2021ErdosProblems} & 1,000+ & Research & \greencheck & \redcross & \redcross & \redcross \\
IMO-AnswerBench \cite{luong2025towards} & 400 & Olympiad & \redcross & \redcross & \greencheck & \greencheck \\
Opt.\ Constants \cite{optimization-constants-repo} & 96 & Research & \greencheck & \greencheck & \redcross & \redcross \\
\specialrule{\lightrulewidth}{\aboverulesep}{0pt}
\rowcolor{light-blue} \rule[-0.15em]{0pt}{1.2em}\horizonmath & 113 & Research & \greencheck & \greencheck & \greencheck & \greencheck \\
\specialrule{\heavyrulewidth}{0pt}{\belowbottomsep}
\end{tabular}%
}
\end{table}

\subsection{Research-level benchmarks}
\label{sec:related-research}

In response to the rapidly-advancing capabilities of AI research agents and the saturation of existing benchmarks, several research-level benchmarks have recently appeared. These benchmarks require novel solutions to unsolved problems rather than reproductions of known answers and are typically hand-crafted by experts.

FrontierMath~\cite{glazer2024frontiermath} comprises several hundred unpublished mathematics problems spanning difficulty tiers from undergraduate through research level. Its most advanced tier targets research-level mathematics, and its ``Open Problems'' subset contains 14 unsolved problems, some of which have already been solved and removed from the dataset \cite{epoch2026frontiermath_open}. Moreover, their verifiers are all private; proposed solutions can only be checked by purchasing access to the verifier. IMProofBench \cite{schmitt2025improofbench} also offers a set of 39 research-level proof problems written by human experts. Both keep their problems and verifiers private or rely heavily on expert human grading, limiting reproducibility.

Similarly, First Proof~\cite{abouzaid2026first} comprises 10 contamination-resistant questions drawn from its authors' unpublished research, but its small scale, manual evaluation, and proof orientation limit its scalability for fast and rigorous model evaluation.

\subsection{AI-driven mathematical discovery}
\label{sec:related-discovery}

One of the first LLM-driven results on open problems came from FunSearch~\cite{romera2024mathematical}, which evolved programs to construct large cap sets beating the best known bounds in extremal combinatorics and discover improved heuristics for bin-packing. AlphaEvolve~\cite{novikov2025alphaevolve} generalized this evolutionary approach and made improvements to complex matrix multiplication algorithms and a kissing-number lower bound. AlphaEvolve's results targeted many topics outside of mathematics, so Georgiev, Tao, G{\'o}mez-Serrano, and Wagner~\cite{georgiev2025mathematical} applied AlphaEvolve to 67 problems across analysis, combinatorics, geometry, and number theory, rediscovering most best-known constructions and improving several---such as new bounds for the finite field Kakeya problem and an optimal tile arrangement for a problem from the 2025 International Mathematical Olympiad.

In parallel, human-AI collaboration has resolved further open problems: Woodruff et al.~\cite{woodruff2026accelerating} used Gemini Deep Think on problems in theoretical computer science, information theory, and physics, and Knuth~\cite{knuth2026claude} similarly used Claude Opus 4.6 to obtain a closed-form Hamiltonian-cycle decomposition of a Cayley digraph---a longstanding open problem.

Recent systems push even further toward autonomy. DeepMind's Aletheia~\cite{feng2026semi} produced multiple papers with minimal human intervention. OpenAI's GPT-5.2 through 5.6 Pro models have similarly resolved many Erd\H{o}s problems~\cite{sothanaphan2026resolution, alexeev2026primitivesetsvonmangoldt}. In physics, GPT-5.2 Pro conjectured a closed-form expression for single-minus gluon amplitudes~\cite{guevara2026single} which was then verified by the human authors. Since mid-2026, this trend has accelerated further. Anthropic's Claude Fable 5 produced an explicit counterexample to the Jacobian conjecture in dimension three~\cite{alpoge2026jacobian}. OpenAI's internal Astra model produced ten results resolving or advancing longstanding problems, including the existence of non-sofic groups, a disproof of Connes's rigidity conjecture, and three Erd\H{o}s problems~\cite{openai2026tenadvances}. An unreleased Claude model also proved that more than two thirds of the nontrivial zeros of the Riemann zeta function are simple and lie on the critical line~\cite{anthropic2026riemann}. This rapid growth in research problems solved by AI motivates a standardized and scalable supply of unsolved yet meaningful research problems that can be used to efficiently evaluate such systems.

\subsection{Related domains and datasets}
While not strictly mathematics, FrontierCS~\cite{mang2025frontiercsevolvingchallengesevolving} shares our design principles: it provides 100+ open-ended problems across algorithmic optimization and computer science. Each problem is hand-crafted, but none have a known optimal solution, and any proposed solution can be deterministically scored on a 0-100 scale by an automated evaluator. FrontierCS provides expert-authored reference solutions, reproducible evaluation harnesses, and dynamic difficulty scaling. Though not mathematics-focused, its open-ended design, automatic verification, and unsolved nature inspire our work.

Beyond formal benchmarks, curated databases of open problems have also become proving grounds for AI. Bloom's Erd\H{o}s Problems site~\cite{Bloom2021ErdosProblems} maintains a collection of Erdős problems, and the Tao-Ivanisvili-Davis optimization-constants repository lists problems and known bounds but provides no evaluation harness. Finally, another related direction includes the UQ (Unsolved Questions) benchmark~\cite{nie2025uq}, which evaluates models on unsolved questions sourced from Stack Exchange. However, practical issues such as question ambiguity, verifier brittleness, and the need for community involvement limit its reliability as a true measure of AI progress on unsolved problems.


\definecolor{hmblue}{HTML}{2F6DB3}
\definecolor{hmbluefill}{HTML}{EDF4FC}
\definecolor{hmgreen}{HTML}{2F8F6B}
\definecolor{hmgreenfill}{HTML}{EEF8F3}
\definecolor{hmgold}{HTML}{C69214}
\definecolor{hmgoldfill}{HTML}{FFF6DB}

\section{Benchmark design and evaluation}
\label{sec:design}

\subsection{Design principles}
\label{sec:principles}

\horizonmath follows four design principles. First, each problem must require an \emph{explicit answer} in the form of a definite mathematical object such as a number, polynomial, set, or graph, rather than a natural-language proof. Second, this answer must be \emph{objectively verifiable} by a simple deterministic computational procedure, whether by numerical comparison against a high-precision reference value, by checking improvement over a published baseline, or by validating that a construction satisfies all required properties. Third, the problem must demand long-horizon \emph{mathematical reasoning} and resist solution by standard computational algorithms, such as computer algebra systems, numerical optimization, or naive searches. Fourth, each problem must have \emph{scientific significance}, meaning it is sourced from the mathematical literature or active research rather than artificially constructed for the benchmark.

All problems are sourced from the research literature (such as journal and conference publications, textbooks, and arXiv preprints) across three categories: problems without known closed-form solutions, optimization problems whose current results are not proven optimal, and construction problems where the target object has not been found. AI agents---specifically, Codex and Claude Code---with web search were used to compile a large, initial list of relevant problems, which was iteratively refined to this subset by human experts. Importantly, this made the collection process significantly cheaper and more efficient than problems in comparable benchmarks, which were all compiled and produced by human experts. Furthermore, none of the problems in the unsolved categories of the benchmark have a known exact or optimal solution in the required form. 

\subsection{Dataset composition}
\label{sec:taxonomy}

We classify each problem along three axes: the structural output type of the expected solution, the solvability level, and the mathematical domain. 

\paragraph{Output types:} All problems require concrete solutions formulated in Python code rather than natural-language proofs. The majority are constant discovery problems, where the model must identify an exact closed-form expression for a numerically approximated target value. Construction tasks form the second largest category, requiring the model to produce discrete mathematical objects---such as lattices, point configurations, or adjacency matrices---that improve upon existing baselines. The remaining problems involve generating computable functions with specific properties or discovering general formulas.

\paragraph{Solvability levels:} We define four levels to gauge difficulty. Level~0 problems are problems that have already been solved by humans (meaning there exist known closed forms or constructions), included for calibration. This does not imply that all models are capable of solving these problems, however. Levels 1–3 are problems currently unsolved by humans and are assigned the following tiers based on the expertise of the authors. Level~1 problems are judged to be likely solvable with known techniques or near-term capabilities; Level~2 problems require significant new insights and are more challenging than Level~1 problems; and Level~3 problems are problems whose potential solutions would be considered major breakthroughs. Since these assignments were made by authors of the benchmark who have mathematical experience in the relevant subfields, they are not objective and are intended as a guide rather than a precise difficulty score.

\paragraph{Mathematical domains:}
The benchmark contains problems across eight subject areas: \emph{Number Theory}, \emph{Special Functions}, \emph{Statistical Mechanics}, \emph{Discrete Geometry}, \emph{Combinatorics}, \emph{Spectral Theory}, \emph{Continuum Physics}, and \emph{Coding Theory}. Each problem is assigned to its corresponding mathematical subject. The distribution in Figure~\ref{fig:benchmark-composition} reflects the current snapshot, but since it is public and welcomes new contributions from the community, its composition will evolve.

\subsection{Admissible solutions for closed forms}
\label{sec:admissible}

For closed-form discovery problems, we enforce strict structural requirements. While the definition of a closed-form is somewhat field-dependent, we adopt a fairly open-ended convention: it must be a finite symbolic expression built exclusively from the following set of operations: rationals, algebraic numbers, standard transcendentals ($\pi$, $e$, Euler's $\gamma$, Catalan's $G$), elementary functions ($\exp$, $\log$, trigonometric, and hyperbolic functions), and special functions as long as they are defined in \texttt{mpmath} \cite{borwein2013closed, johansson2013mpmath}. We choose this flexible definition given the lack of a universal agreement on what "closed-form" means. 

We explicitly forbid solutions that include unevaluated integrals, infinite series, limits, implicit definitions, and numerical approximations presented as exact values---except for those that are special functions. We also do not permit solutions that use computational tools such as numerical integration, truncated infinite series, numerical root-finding, and computation of resultants by evaluating one polynomial at the roots of another. We also exclude problem classes that can be solved via standard deterministic algorithms, such as the Remez algorithm, PSLQ, gradient descent, or symbolic integration. Finally, we discourage symbolic parameter fitting to the ground-truth numerical solution by only showing at most five significant figures of precision in the prompt.


\section{Automated verification}
\label{sec:infrastructure}
A core feature of our benchmark is a fully automated and reproducible evaluation pipeline. Unlike theorem-proving benchmarks, which require complex formalization in proof assistants or expert grading, and unlike other verifiable benchmarks that do not release their evaluation infrastructure, our framework enables fast, automatic verification.

\subsection{Execution pipeline and admissibility}
\label{sec:execution_admissibility}
Models are prompted to generate a self-contained Python function, \texttt{def proposed\_solution()}, returning the solution as either an \texttt{mpmath} symbolic expression or a JSON-serializable dictionary representing a discrete construction. Strict admissibility criteria are enforced on the proposed solutions in the system prompt as well as via the compliance checker. Closed-form expressions must be finite symbolic combinations of permissible operations (rational numbers, elementary functions, standard transcendentals at algebraic arguments); infinite series, limits, implicit definitions, and numerical approximations are forbidden. 
Explicit constructions must follow the output type defined in the function template. These constraints are enforced via an LLM-based compliance checker, which inspects the proposed solution for forbidden operations such as numerical root-finding, quadrature, and hard-coded constants.

\subsection{Evaluation modes and scoring mechanics}
\label{sec:eval_modes}

\begin{figure}
    \centering
\includegraphics[width=1.0\linewidth]
{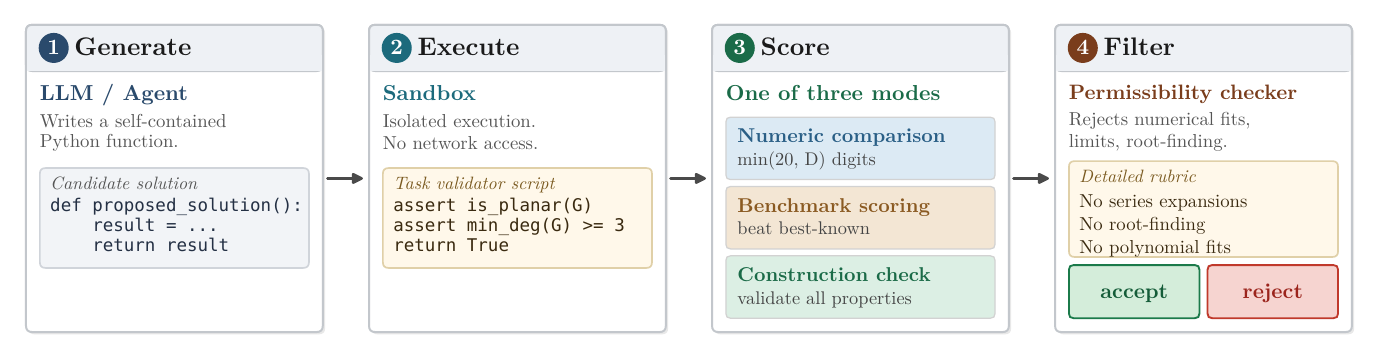}
\caption{\horizonmath's automated evaluation pipeline. A model generates a proposed solution, which is screened by a compliance checker that rejects forbidden operations. Admissible solutions are routed to one of three evaluation modes: numeric comparison against a ground-truth value, benchmark scoring for improvement over the best-known result, and construction checking to validate all required structural properties.}
    \label{fig:pipeline}
\end{figure}

\textbf{Closed-form} problems require the model to produce a symbolic closed-form expression, evaluated to high precision via \texttt{mpmath} and compared against a reference value typically computed via integration or series expansion. A candidate is accepted if it matches the reference to $\min(20, D)$ decimal digits, where $D$ is the number of verified digits available for that ground truth. \textbf{Bound-improvement} problems target optimization tasks where the global optimum is unknown but a published baseline exists. The model's output is executed and scored by problem-specific validators on the relevant metric (e.g., packing density, asymptotic bounds). A solution passes if it strictly improves upon the best-known result, and solutions are further ranked by relative improvement over the baseline. \textbf{Existence} problems concern construction questions, such as finding a Hadamard matrix of an unresolved order or a set of mutually orthogonal Latin squares of a given size. Scoring is pass/fail: deterministic, problem-specific validators exhaustively check whether the proposed object satisfies all required properties, with no baseline comparison.\footnote{In the released dataset these modes correspond to the identifiers \texttt{ground\_truth\_computable}, \texttt{benchmark\_best\_known}, and \texttt{new\_construction}, respectively.}

We acknowledge that while matching a high-precision reference value does not formally prove a closed-form expression is exactly correct, \emph{such solutions that pass the permissibility check can be regarded as high-confidence conjectures until proven}. While a formal proof would be needed to guarantee the correctness, this numerical check remains essential to producing novel mathematical research, since new theorems often first begin with highly-accurate conjectures. Constructions that beat a published bound or satisfy the corresponding validator, by contrast, are verified deterministically by checking all required properties. 


\section{Representative problem examples}
\label{sec:examples}

\enlargethispage{2\baselineskip}

We illustrate the benchmark with three representative problems covering the \texttt{closed-form}, \texttt{bound-improvement}, and \texttt{new-construction} categories; the examples below show the core task and output specification given to the model. 

\begin{compactprompt}{1.\; Spinor Norm Elliptic Integral ($I_0$)}
\textit{Special Functions} \,$\mid$\, \textit{Constant} \,$\mid$\, \texttt{Closed-Form} \,$\mid$\, \textit{Solvability 1}

\smallskip
\textbf{Task.}\; Let $K(m)$ and $E(m)$ denote the complete elliptic integrals of the first and second kinds in the parameter convention. Define
\[
I_0=\frac{4\sqrt{2}}{\pi}\int_0^1
\frac{1+\sqrt{z}}{(1+z)^3}
\left(2E(z)-(1-z)K(z)\right)\,dz.
\]
This elliptic-integral constant arises in a spinor norm calculation and has numerical value
$I_0\approx 1.77425\dots$, but no closed-form expression is given in~\cite{takahashi2024basis}. Find a closed-form symbolic expression for $I_0$.

\smallskip
\textbf{Output format.}\; An expression using only mpmath-permitted functions. No quadrature, infinite series, root-finding, or hard-coded numerical constants. (Exact output format specified in Appendix \ref{spinor_norm}.)
\end{compactprompt}

\begin{compactprompt}{2.\; Minimum-Scope Difference Triangle Set $(7,5)$}
\textit{Combinatorics \& Design Theory} \,$\mid$\, \textit{Construction} \,$\mid$\, \texttt{Bound-Improvement} \,$\mid$\, \textit{Solvability 1}

\smallskip
\textbf{Task.}\; An $(n,k)$-DTS is an $n\times(k+1)$ integer array $A$ with rows strictly increasing from $0$ whose positive within-row differences $\{a_{i,j}-a_{i,j'}:j'<j\}$ are pairwise distinct across all rows; the \emph{scope} is $m(A)=\max_{i,j}a_{i,j}$, with applications in coding and communications~\cite{shehadeh2026new}. Find a valid $(7,5)$-DTS with scope strictly below the published bound $m(7,5)\le 112$, i.e.\ $m(A)\le 111$.

\textbf{Output format.}\; A dictionary with integer entries and strictly increasing rows. The validator computes the scope and verifies pairwise distinct differences exactly.
\end{compactprompt}

\begin{compactprompt}{3.\; Three Mutually Orthogonal Latin Squares of Order 10}
\textit{Coding Theory} \,$\mid$\, \textit{Construction} \,$\mid$\, \texttt{New-Construction} \,$\mid$\, \textit{Solvability 1}

\smallskip
\textbf{Task.}\; A Latin square of order $10$ is a $10\times 10$ array over $\{0,1,\ldots,9\}$ in which every symbol occurs exactly once in each row and column. Two such squares are \emph{orthogonal} if their superposition contains each of the $100$ ordered pairs exactly once. Whether three mutually orthogonal Latin squares of order $10$ exist remains open~\cite{rubin2021integerconstraintprogrammingrevisited}. Construct three Latin squares $L_1,L_2,L_3$ of order $10$ such that each of $(L_1,L_2)$, $(L_1,L_3)$, and $(L_2,L_3)$ is orthogonal.

\textbf{Output format.}\; A dictionary \texttt{\{"squares": [L1, L2, L3]\}} containing three $10\times 10$ integer arrays with entries in $\{0,\ldots,9\}$. The validator checks the Latin-square conditions and all three pairwise orthogonality conditions exactly.
\end{compactprompt}

\section{Evaluation results}
\label{sec:Results}

\begin{figure}
    \centering
    \includegraphics[width=1.0\linewidth]{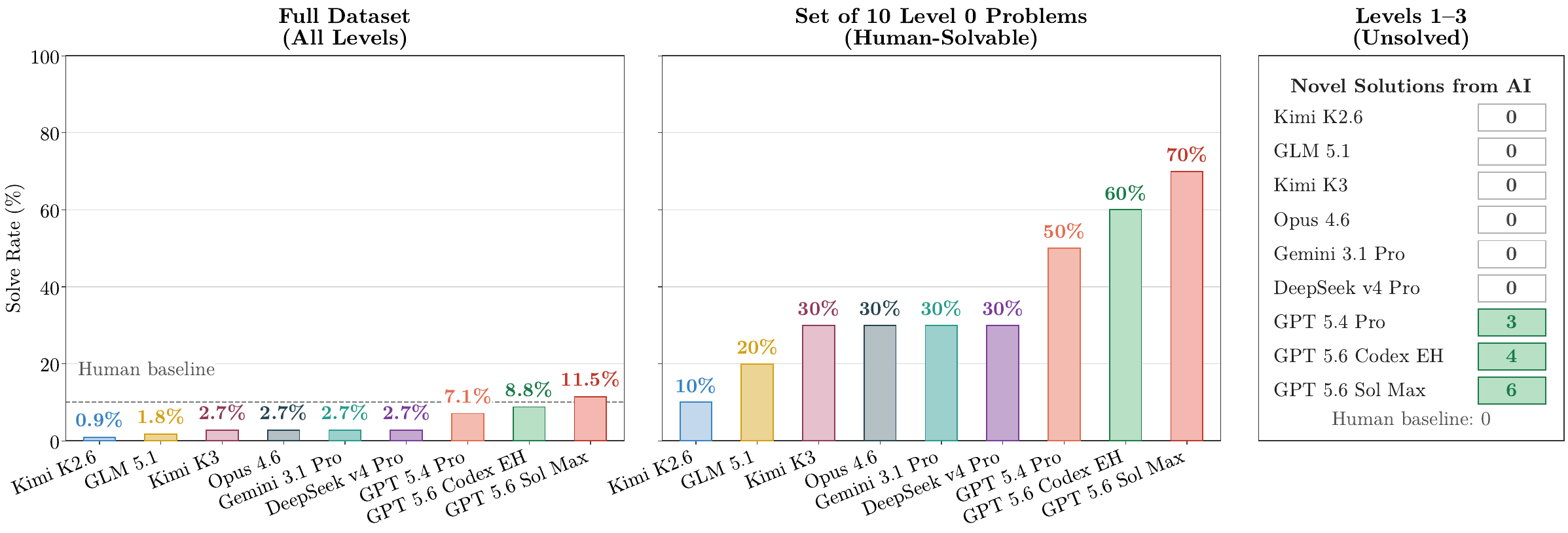}
\caption{Model performance on \horizonmath across nine frontier models: Kimi K2.6, GLM 5.1, DeepSeek V4 Pro, Claude Opus 4.6, Gemini 3.1 Pro, GPT 5.4 Pro, GPT 5.6 Sol xhigh in Codex, and GPT 5.6 Sol Max. The dashed line (human baseline) indicates Level 0 problems with known solutions. \emph{Left:} Solve rate on the full benchmark (113 problems). GPT 5.6 Sol Max leads at 11.5\% (13 of 113), comprising 7 problems from the calibration tier (solvability~0) and 6 from the unsolved tiers (solvability~1--3). \emph{Center:} Solve rate restricted to the ten solvability-0 calibration problems, all of which are solvable by humans. \emph{Right:} Number of admissible candidate solutions on the unsolved tiers (solvability~1--3). Three models produce novel solutions. We discuss these results further in Section \ref{sec:Results} and Appendix \ref{sec:Appendix}.}
\label{fig:placeholder}
\end{figure}

We evaluate nine frontier models on HorizonMath: five closed models (\textbf{Claude Opus 4.6}, \textbf{Gemini 3.1 Pro}, \textbf{GPT 5.4 Pro}), \textbf{GPT 5.6 Sol xhigh in Codex}, and \textbf{GPT 5.6 Sol Max}, and four open-weight models (\textbf{Kimi K2.6 \& K3}, \textbf{GLM 5.1}, and \textbf{DeepSeek V4 Pro}). All models outside of GPT 5.6 Sol xhigh in Codex are set to the highest reasoning effort with the maximum output-token limit. The open-weight models are run with maximum reasoning effort and a 128k token limit. All evaluations are pass@1 due to the high costs of running frontier models on problems of this difficulty, except on questions where GPT 5.4 Pro produced a verified solution on its first attempt, where we ran one additional query with GPT 5.4 Pro to see whether we could obtain a larger improvement over the baseline.

On solvability~1--3 problems, the only models that make important progress are \textbf{GPT 5.4 Pro}, \textbf{GPT 5.6 xhigh in Codex}, and \textbf{GPT 5.6 Sol Max}. GPT 5.4 Pro proposes novel solutions to three problems in the solvability~1 tier: the \emph{Thin-Triangle Kakeya (128 slopes)} and \emph{Asymptotic Upper Bound Constant for Diagonal Ramsey Numbers} optimization problems, where its constructions improve on the best-known published baselines, and the \emph{Spinor Norm Integral ($I_0$)} closed-form problem, where it produces a symbolic expression matching the high-precision numerical reference. GPT 5.6 Sol Max produces all three of these results and novel solutions to three additional problems in the solvability~2 tier: the \emph{Fifth Moment of the Airy Function ($a_5$)}, the \emph{Equal-Power TE+TM Spherical-Mode Quality Factor}, and the \emph{Non-Resonant TM/TE Spherical-Mode Quality Factor}. Its symbolic expressions for these three additional problems also pass the compliance checks and match the corresponding high-precision numerical references. All six problems and proposed solutions have been verified by domain experts and are detailed in Appendix~\ref{sec:Appendix}. No other evaluated closed model (Claude Opus 4.6 or Gemini 3.1 Pro) or open-weight model (Kimi K2.6, GLM 5.1, or DeepSeek V4 Pro) produced candidate solutions that passed both the compliance checker and the verifiers on the unsolved tiers.

On the 10 solvability-0 calibration problems, Kimi K2.6 solves 1, GLM 5.1 solves 2, and Kimi K3 and DeepSeek V4 Pro solve 3. Among the closed models, Claude Opus 4.6 and Gemini 3.1 Pro each solve 3, GPT 5.4 Pro correctly solves 5, GPT 5.6 Sol xhigh in Codex solves 6, and GPT 5.6 Sol Max solves 7. We analyze these results in Section \ref{sec:reasoning_appendix}.

\section{Analysis}
\label{sec:analysis}
Here, we study the reasoning behavior across models. We find notable differences in token and cost efficiency, and compare open-source models against closed-source ones.

\subsection{Model reasoning and behavior}

\begin{figure}[htbp]
    \centering
    \begin{subfigure}[t]{0.5\textwidth}
        \includegraphics[width=\textwidth, trim=0 0 75 0, clip]{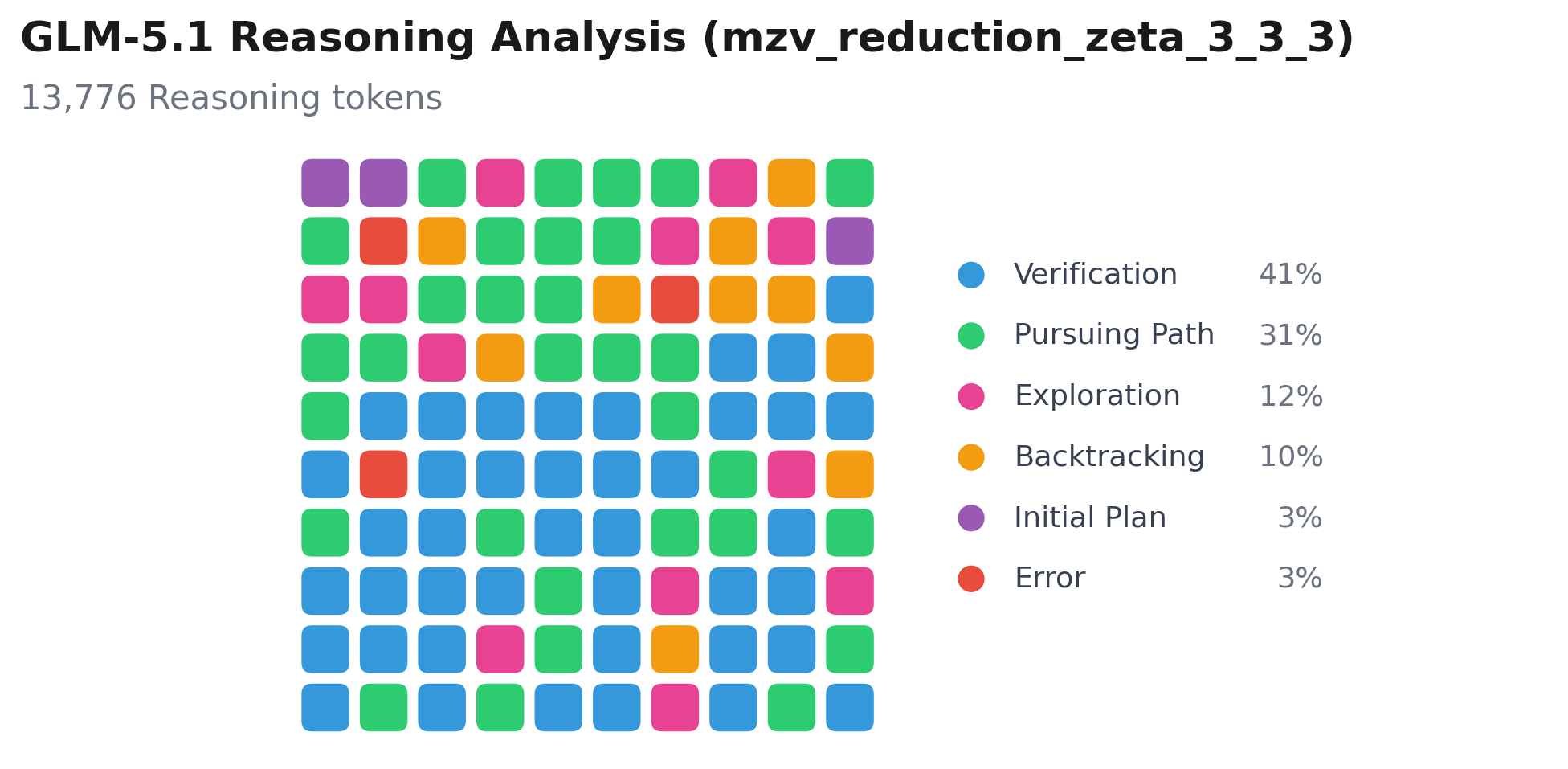}
        \caption{A correct response to a solvability-0 problem.}
    \end{subfigure}%
    \begin{subfigure}[t]{0.5\textwidth}
        \includegraphics[width=\textwidth, trim=0 0 75 0, clip]{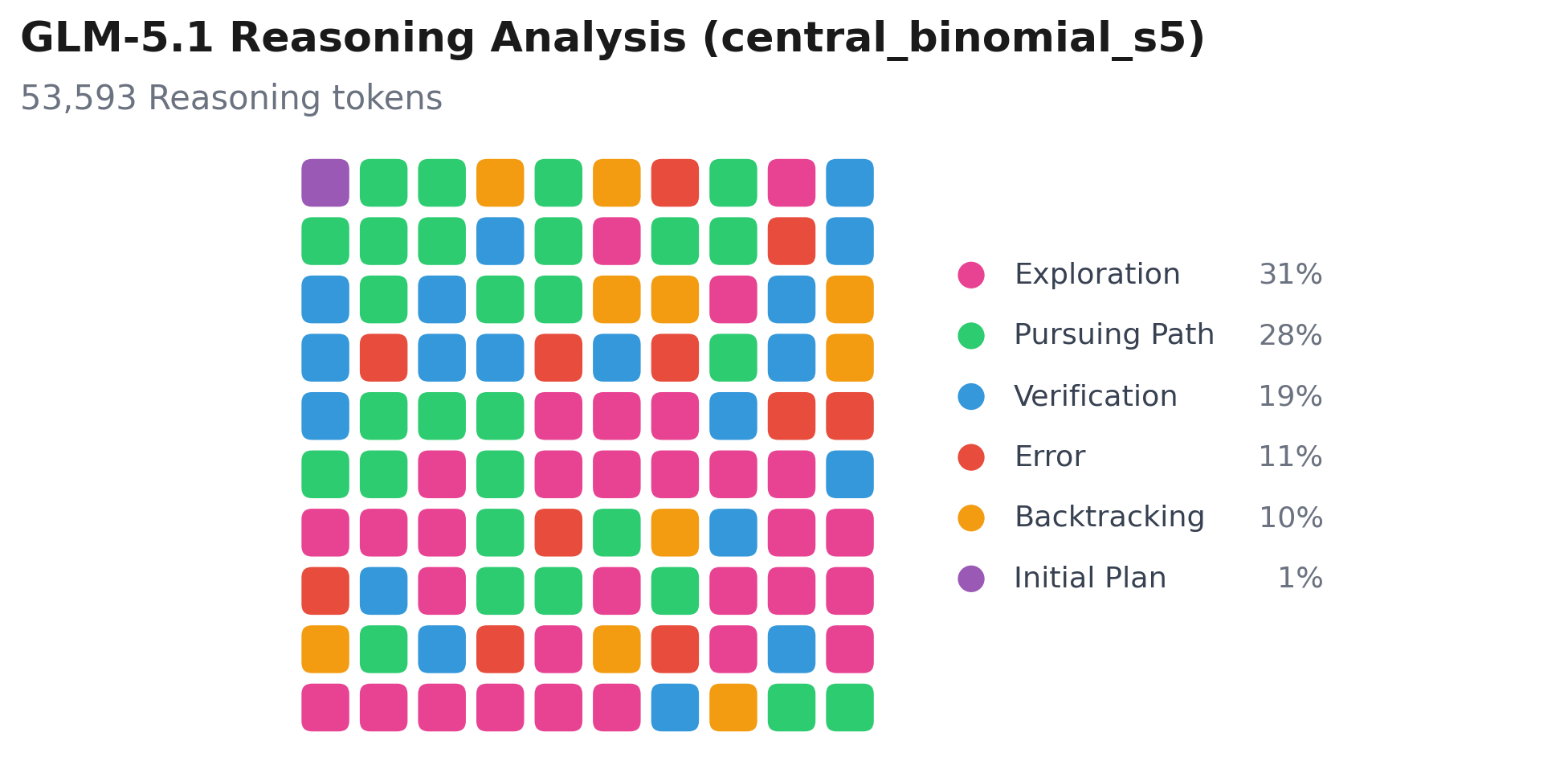}
        \caption{An incorrect response to a solvability-0 problem.}
    \end{subfigure}
    \caption{Reasoning trace analysis for GLM 5.1. A correct response to a solvable problem uses fewer tokens compared to an incorrect response to a similarly difficult problem, which involves significantly more exploration attempts and errors along with less verification. Read left-to-right, top-to-bottom; each square represents 1\% of the model's reasoning tokens for that problem.}
    \label{fig:comparison}
\end{figure}

In Figure~\ref{fig:comparison}, we contrast the reasoning traces of GLM 5.1 on two solvability-0 problems: a correct response uses 13,776 tokens, while an incorrect one on a similarly difficult problem uses 53,593, nearly four times as many. This pattern holds across models, with average tokens on incorrect runs uniformly exceeding those on correct ones (\textbf{24.7k vs 63.0k} for DeepSeek V4 Pro, \textbf{18.2k vs 62.1k} for GLM 5.1, \textbf{43.5k vs 59.2k} for Kimi K2.6, \textbf{16.3k vs 23.1k} for Gemini 3.1 Pro, \textbf{43.8k vs 82.4k} for Claude Opus 4.6, \textbf{30.1k vs 40.3k} for GPT 5.4 Pro, and \textbf{18.4k vs 39.9k} for GPT 5.6 Sol Max). In Appendix~\ref{sec:reasoning_appendix}, we measure the rate at which open-weight models switch reasoning strategies. We find that this rate increases with problem difficulty within each model, and that weaker models switch more often than stronger ones.

The composition of reasoning tokens also looks different between correct and incorrect runs. We classify each segment of the open-weight models’ traces into one of six categories (verification, pursuing/solving a path, backtracking, initial planning, exploration, or finding errors) and observe that correct responses spend a greater share of tokens on planning and verification, while incorrect ones spend more on exploration. This is consistent with the pattern observed above, where incorrect runs use more tokens overall by trying more exploration strategies. We discuss reasoning behaviors, especially related to models giving up early, in Appendix \ref{app:qualitative}.

Finally, we count solutions that pass naive numerical verification but are flagged by the compliance checker for operations not permitted by the admissibility criteria. GPT 5.6 Sol Extra High (Codex) produces the most non-compliant solutions (50), followed by GPT 5.4 Pro (43), GPT 5.6 Sol Max (16), Claude Opus 4.6 (13), Gemini 3.1 Pro (10), DeepSeek V4 Pro (6), GLM 5.1 (5), and Kimi K2.6 (4). The GPT models' high counts are consistent with their higher overall attempt rates. These rates illustrate why compliance checking is necessary: without it, numerical agreement alone would significantly overstate model capability.

\subsection{Token and cost efficiency}

\begin{figure}[htbp]
    \centering
    \begin{minipage}[c]{0.45\textwidth}
        \centering
        \includegraphics[width=\linewidth]{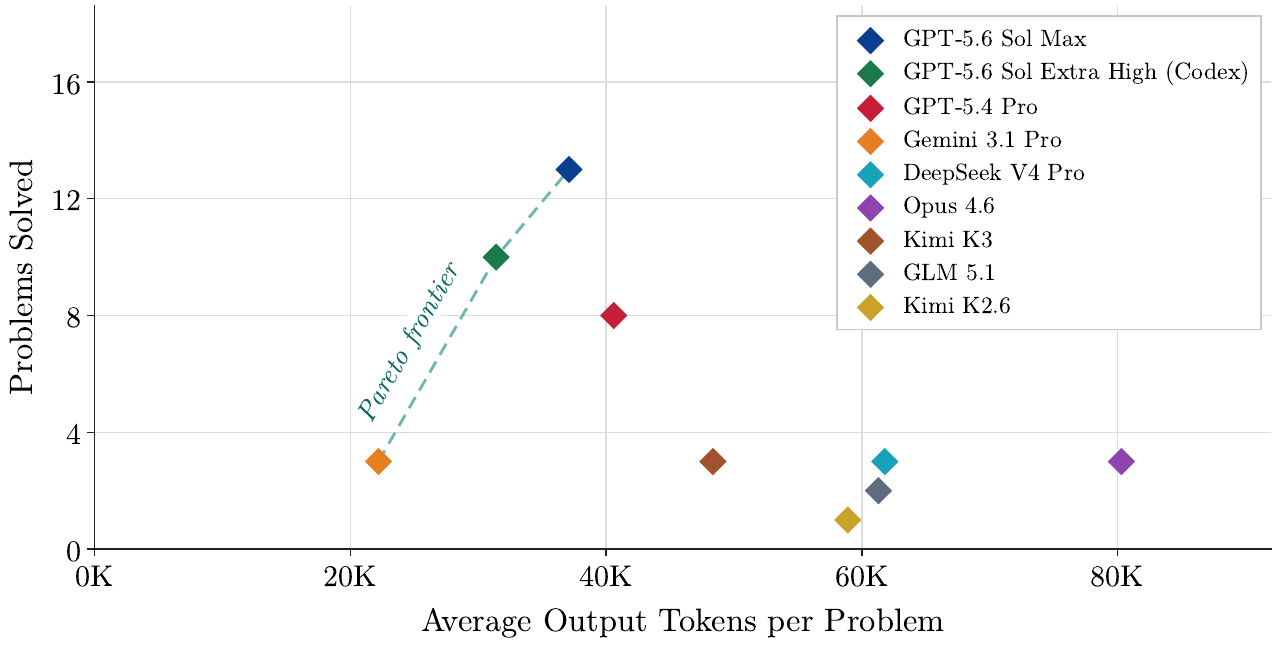}
        \caption{The Pareto frontier of token efficiency versus accuracy across models.}
        \label{fig:pareto}
    \end{minipage}
    \hspace{.4em}
    \begin{minipage}[c]{0.5\textwidth}
    \centering
    \resizebox{\linewidth}{!}{%
    \begin{tabular}{lcc}
    \toprule
    \textbf{Model} & \textbf{Avg. Output Toks} & \textbf{Toks / Solve} \\
    \midrule
    GPT-5.6 Sol Max               & 37.1k & \textbf{0.322} \\
    GPT-5.6 Sol xhigh (Codex)     & 31.4k & 0.355 \\
    GPT-5.4 Pro                   & 40.6k & 0.573 \\
    Gemini 3.1 Pro                & 22.2k & 0.836 \\
    Kimi K3                       & 48.4k & 1.821 \\
    DeepSeek V4 Pro               & 61.8k & 2.328 \\
    Opus 4.6                      & 80.3k & 3.025 \\
    GLM 5.1                       & 61.3k & 3.463 \\
    Kimi K2.6                     & 58.9k & 6.656 \\
    \bottomrule
    \end{tabular}%
    }
    \vspace{0.2cm}
    \captionof{table}{Analysis of model efficiency across output tokens, ordered by tokens per solved problem. Toks / Solve is the number of output tokens, in millions, per solved problem; lower is better.}
    \label{tab:model_efficiency}
\end{minipage}
        
\end{figure}

Figure~\ref{fig:pareto} and Table~\ref{tab:model_efficiency} characterize the accuracy-cost tradeoff across the six models. The Pareto frontier of solved problems versus average output tokens is defined by three closed models, Gemini 3.1 Pro (3 problems at 22.2k tokens on average), GPT 5.6 xhigh in Codex (10 at 31.4k), and GPT 5.6 Sol Max (15 at 37.1k). Token efficiency (solved problems per 1M tokens) follows the same ordering: GPT 5.4 Pro leads at 1.744 and Gemini 3.1 Pro at 1.196, while the open-weight models fall between 0.150 and 0.430.

\section{Discussion}
We introduced \horizonmath, a benchmark of over 100 mathematical problems from several subdomains of mathematics, physics, and computer science that are predominantly unsolved and sourced from the research literature. Each problem outside of the calibration set lacks a known solution but can automatically be verified with high probability, drastically lowering the possibility of data contamination and accelerating progress for the research community. The evaluation framework is publicly available and computational, avoiding the scalability bottlenecks of formal proof verification and decreasing the need for expert review.

We acknowledge key limitations. Matching a high-precision numerical reference, even to 20 decimal digits, does not formally prove that a closed-form expression is exactly correct; such solutions are best regarded as strong conjectures until proven. Our compliance checker, which uses an LLM to detect forbidden operations in proposed solutions, is similarly imperfect: it may occasionally accept solutions that exploit subtle loopholes or reject valid ones that use unusual but legitimate constructions. HorizonMath therefore does not entirely replace expert verification, but substantially reduces the human effort required by automatically filtering many incorrect or inadmissible solutions.

We see two complementary directions for extending HorizonMath: (1) expanding what counts as an admissible solution, and (2) expanding the class of problems themselves. The first direction could involve accepting solutions that are simplifications but not necessarily exact closed forms according to the definition used in this benchmark. This flexibility would help capture a fuller spectrum of research, especially in fields like physics, where the goal is not always an exact closed-form solution but a reduction in complexity that increases computational tractability. The second direction would expand beyond the existing problem categories to include open problems requiring proof-based verification, by integrating with formal verification systems such as Lean. This would require carefully producing theorem statements in Lean to begin with, but would enable HorizonMath to encompass a broader class of unsolved problems, including those where the contribution is a proof rather than a concrete object, while maintaining our automated and reproducible evaluation strategy. Overall, HorizonMath benchmarks frontier models on unsolved research problems with automated verification, and improvements on it reflect real advances in mathematical reasoning capabilities.

\bibliographystyle{plain}
\bibliography{bib}

\appendix
\newpage

\definecolor{promptbg}{RGB}{245,245,252}
\definecolor{promptframe}{RGB}{100,100,180}
\definecolor{codebg}{RGB}{248,248,248}
\definecolor{codeframe}{RGB}{180,180,180}
\definecolor{solnbg}{RGB}{240,252,240}
\definecolor{solnframe}{RGB}{60,140,60}
\definecolor{resultbg}{RGB}{255,248,235}
\definecolor{resultframe}{RGB}{200,150,50}

{\noindent\Large\textbf{Appendix}\par}

\section{Detailed problem statements and solutions}
\label{sec:Appendix}

We detail three problems of the \texttt{solvability 1} tier for which GPT-5.4 Pro produced a novel closed-form solution or an optimization that beat the best known baseline.

\newtcolorbox{solnbox}{
  colback=solnbg, colframe=solnframe,
  boxrule=0.6pt, arc=3pt, left=8pt, right=8pt, top=6pt, bottom=6pt,
  title={\textbf{Solution Output \& Comparison}},
  fonttitle=\normalsize, coltitle=black, attach boxed title to top left={yshift=-2mm,xshift=4mm},
  boxed title style={colback=solnbg, colframe=solnframe, boxrule=0.4pt}, breakable
}

\newtcolorbox{resultbox}{
  colback=resultbg, colframe=resultframe,
  boxrule=0.6pt, arc=3pt, left=8pt, right=8pt, top=6pt, bottom=6pt, breakable
}

\lstset{
  basicstyle=\small\ttfamily,
  backgroundcolor=\color{codebg},
  frame=single,
  rulecolor=\color{codeframe},
  breaklines=true,
  columns=fullflexible,
  keepspaces=true,
  tabsize=4,
  showstringspaces=false,
  language=Python,
  keywordstyle=\color{blue!70!black},
  commentstyle=\color{green!50!black},
  stringstyle=\color{red!60!black},
}

\setlength{\parindent}{0pt}


\subsection{Spinor Norm Elliptic Integral}
\label{spinor_norm}

\vspace{0.3em}

\textbf{Domain:} Special Functions $\mid$ \textbf{Output Type:} Constant \\
\textbf{Evaluation Mode:} \texttt{ground\_truth\_computable} $\mid$ \textbf{Solvability:} 1 \\
\textbf{Background:} This benchmark asks for a symbolic closed form for an elliptic-integral constant arising in a spinor norm calculation. The integral representation involves complete elliptic integrals of the first and second kinds, and the source paper derives the representation but does not provide a closed form. \\
\textbf{Source:}
Takahashi (2024), \textit{Basis of spinors expressed by differential forms and calculating its norm}, arXiv:2403.13003.

\vspace{0.5em}

\begin{promptbox}

Consider the following elliptic-integral expression arising in a spinor norm calculation.

\vspace{0.5em}

Let $K(m)$ and $E(m)$ denote the complete elliptic integrals of the first and second kinds with parameter $m$. Define
\[
I_0=
\frac{4\sqrt{2}}{\pi}
\int_0^1
\frac{1+\sqrt{z}}{(1+z)^3}
\left(2E(z)-(1-z)K(z)\right)\,dz.
\]

\vspace{0.3em}

\textbf{Task:} Find a closed-form symbolic expression for the constant $I_0$.

\vspace{0.3em}

\textbf{Inadmissible approaches:}
\begin{itemize}
  \item Numerical quadrature or numerical integration routines such as \texttt{mp.quad}.
  \item Returning this integral, an equivalent transformed integral, or a custom integral transform.
  \item Hardcoding the decimal value or using a tuned approximation.
  \item Infinite series or truncations whose accuracy depends on the number of terms.
\end{itemize}

\vspace{0.3em}

\rule{\linewidth}{0.4pt}

\vspace{0.3em}

\textbf{REQUIRED OUTPUT FORMAT:}

\begin{tcolorbox}[colback=codebg, colframe=codeframe,
  boxrule=0.4pt, arc=2pt, left=6pt, right=6pt, top=4pt, bottom=4pt, breakable]
\begin{lstlisting}
def proposed_solution():
    from mpmath import mp
    mp.dps = 100
    result = ...
    return result
\end{lstlisting}
\end{tcolorbox}

\end{promptbox}


\begin{tcolorbox}[
  colback=codebg, colframe=codeframe,
  title=\textbf{Generated Code},
  coltitle=black,
  colbacktitle=green!12!white,
  boxrule=0.8pt, arc=2pt,
  left=6pt, right=6pt, top=4pt, bottom=4pt, breakable
]
\begin{lstlisting}
def proposed_solution():
    from mpmath import mp
    mp.dps = 100
    result = (
        mp.gamma(mp.mpf(1)/4)**2 / (8 * mp.sqrt(mp.pi))
        + mp.gamma(mp.mpf(3)/4)**2 / mp.sqrt(mp.pi)
    )
    return result
\end{lstlisting}
\end{tcolorbox}

\begin{resultbox}
\textbf{Closed-form symbolic solution.}

\[
I_0
=
\frac{\Gamma\!\left(\frac14\right)^2}{8\sqrt{\pi}}
+
\frac{\Gamma\!\left(\frac34\right)^2}{\sqrt{\pi}}.
\]

Numerically, $I_0 \approx
1.774250423444665045...$ matches the ground-truth computed formula to 140 digits of precision, as defined in our numerical validation script.
\end{resultbox}

\begin{solnbox}
\textbf{Approach.}
Consider $I_0 = 2E\!\left(\tfrac{1}{2}\right) - \frac{1}{2}\,K\!\left(\tfrac{1}{2}\right).$
Using the self-complementary Legendre relation at $m = \tfrac{1}{2}$,
$$I_0 = \frac{\Gamma\!\left(\tfrac{1}{4}\right)^2}{8\sqrt{\pi}} + \frac{2\pi^{3/2}}{\Gamma\!\left(\tfrac{1}{4}\right)^2}$$
we get this compact derivation:
$$2E(t^2) - (1 - t^2)K(t^2) = (1 + t)\,E\!\left(\frac{4t}{(1+t)^2}\right)$$

(Landen), so with $z = t^2$ and then $u = \frac{4t}{(1+t)^2}$,

$$I_0 = \frac{4\sqrt{2}}{\pi} \int_0^1 \frac{u\,E(u)}{(2-u)^3\sqrt{1-u}}\,du = \frac{8\sqrt{2}}{\pi} \int_0^1 \frac{1 - x^2}{(1 + x^2)^3}\,E(1 - x^2)\,dx.$$

If $A(x) = E(1 - x^2)$, $B(x) = K(1 - x^2)$, then

$$\frac{d}{dx}\frac{x(A - B)}{(1 + x^2)^2} = \frac{2(1 - x^2)A + (3x^2 - 1)B}{(1 + x^2)^3},$$

hence

$$I_0 = \frac{2\sqrt{2}}{\pi}\,F''(1), \qquad F(k) := \int_0^1 \frac{k\,K(1 - x^2)}{k^2 + x^2}\,dx.$$

Now the Beltrami transform plus the imaginary-modulus formula gives

$$F(k) = \frac{\pi}{2\sqrt{1 + k^2}}\,K\!\left(\frac{1}{1 + k^2}\right).$$

Differentiating twice and setting $k = 1$ yields

$$F''(1) = \frac{\pi}{4\sqrt{2}}\left(4E\!\left(\tfrac{1}{2}\right) - K\!\left(\tfrac{1}{2}\right)\right),$$

so

$$I_0 = 2E\!\left(\tfrac{1}{2}\right) - \frac{1}{2}\,K\!\left(\tfrac{1}{2}\right).$$
    
\end{solnbox}

\subsection{Thin-Triangle Kakeya (128 slopes): Minimize Union Area}

\vspace{0.3em}

\textbf{Domain:} Geometry \& Discrete Geometry $\mid$ \textbf{Output Type:} Construction \\
\textbf{Evaluation Mode:} \texttt{benchmark\_best\_known} $\mid$ \textbf{Solvability:} 1 \\
\textbf{Background:} This benchmark is a discrete, thickened Kakeya-type construction in the style of Schoenberg and Keich. The classical Kakeya needle problem asks for the smallest-area planar set containing a unit line segment in every direction; Besicovitch showed such sets can have measure zero. Keich studied a thickened variant with $N$ equally-spaced slopes, where each segment is replaced by a thin triangle of base-width $\delta = 1/N$, and sought to minimize the area of their union. AlphaEvolve (Google DeepMind, 2025) found a construction with union area $\approx 0.11481$, improving on Keich's earlier construction ($\approx 0.11921$). \\
\textbf{Source:}
AlphaEvolve (Google DeepMind, 2025). Baseline from the AlphaEvolve construction; see Novikov et al.\ (2025), \textit{arXiv preprint} arXiv:2506.13131.

\vspace{0.5em}

\begin{promptbox}

Consider the following optimization problem.

\vspace{0.5em}


\textbf{Thin-Triangle Kakeya (128 slopes): Minimize Union Area}

\vspace{0.5em}

\textbf{Definition:} Fix $N = 128$ and $\delta = 1/128$. For each $i = 0, 1, \ldots, 127$, specify a unit line segment
\[
\ell_i = \{(x,\; a_i x + b_i) : x \in [0,1]\} \quad\text{with slope}\quad a_i = i/128.
\]
From each segment $\ell_i$ define the thin triangle $R_\delta(\ell_i)$ as follows:
\begin{itemize}
  \item The \textbf{upper edge} is $\ell_i$.
  \item The \textbf{lower edge} is the segment from $(0,\; b_i - \delta)$ to $(1,\; a_i + b_i)$.
  \item The \textbf{vertical edge} closes the triangle at $x = 0$.
\end{itemize}
Equivalently, for $x \in [0,1]$, the vertical cross-section of $R_\delta(\ell_i)$ is the interval
\[
y \in \bigl[\,a_i x + b_i - \delta(1 - x),\;\; a_i x + b_i\,\bigr].
\]

The output defines the set $E = \bigcup_{i=0}^{127} R_\delta(\ell_i)$.

\vspace{0.3em}

\textbf{Goal:} MINIMIZE $\operatorname{Area}(E)$.

\vspace{0.3em}

\textbf{Current State-of-the-Art:}
\begin{itemize}
  \item \textbf{Metric:} $\operatorname{Area}(E)$
  \item \textbf{Best Known Value:} $\approx 0.11481$ (AlphaEvolve, Google DeepMind, 2025)
  \item \textbf{Direction:} MINIMIZE (lower area is better)
  \item \textbf{Source:} Novikov et al.\ (2025), arXiv:2506.13131
\end{itemize}

\vspace{0.3em}

\rule{\linewidth}{0.4pt}

\vspace{0.3em}

\textbf{Mathematical framework.}\enspace
The slopes are fixed at $a_i = i/128$ for $i = 0, \ldots, 127$. The only free parameters are the 128 intercepts $b_0, b_1, \ldots, b_{127}$. Each thin triangle is the convex hull
\[
R_\delta(\ell_i) = \operatorname{conv}\bigl\{(0,\; b_i - \delta),\; (0,\; b_i),\; (1,\; a_i + b_i)\bigr\}.
\]
The area of the union $E = \bigcup_i R_\delta(\ell_i)$ is computed by exact piecewise-linear integration of the union of cross-sections at each $x$-coordinate.

\vspace{0.3em}

The AlphaEvolve baseline achieves area $\approx 0.1148103258186177$. The AlphaEvolve triangles $\operatorname{conv}\{(x_i, 0),\; (x_i + i/128, 0),\; (x_i + (i{+}1)/128, 1)\}$ map to the formulation above via the area-preserving coordinate swap $(x, y) \to (y, x)$ with $b_i = x_i + i/128$.

\vspace{0.5em}

\textbf{To beat the baseline,} find intercepts giving $\operatorname{Area}(E) < 0.1148103258186177$.

\vspace{0.5em}

\rule{\linewidth}{0.4pt}

\vspace{0.3em}

\textbf{REQUIRED OUTPUT FORMAT:}

\begin{tcolorbox}[colback=codebg, colframe=codeframe,
  boxrule=0.4pt, arc=2pt, left=6pt, right=6pt, top=4pt, bottom=4pt, breakable]
\begin{lstlisting}
def proposed_solution():
    # Must output b_i for each slope i/128.
    return {
        "intercepts": [b_0, b_1, ..., b_127]
    }
\end{lstlisting}
\end{tcolorbox}

\vspace{0.2em}

\begin{itemize}
  \item \texttt{intercepts}: a list of exactly 128 floats $[b_0, b_1, \ldots, b_{127}]$.
  \item Slopes are fixed to $a_i = i/128$.
  \item The validator computes $\operatorname{Area}(E)$ by exact piecewise-linear integration of union cross-sections (deterministic).
\end{itemize}

Return the dictionary.

\end{promptbox}


\vspace{1.5em}

\begin{tcolorbox}[
  colback=codebg, colframe=codeframe,
  title=\textbf{Generated Code},
  coltitle=black,
  colbacktitle=green!12!white,
  boxrule=0.8pt, arc=2pt,
  left=6pt, right=6pt, top=4pt, bottom=4pt,breakable
]
\begin{lstlisting}[
  basicstyle=\small\ttfamily,
  xleftmargin=14pt,
  aboveskip=4pt,
  belowskip=4pt,
]
def proposed_solution():
         N = 128
         INV_Q_UNIT = 1.0 / 8.0   # q-values are stored in eighths
         INV_B_UNIT = 1.0 / 1024.0  # since b_i = q_i / 128, and q_i is in eighths

         # Full Haar-style dyadic basis:
         # parameter (k, s, e) subtracts coeff on indices [s, e),
         # where k is the binary level (higher k = coarser block).
         params = []
         for k in range(6, -1, -1):
             half = 1 << k
             block = half << 1
             for start in range(0, N, block):
                 params.append((k, start + half, start + block))

         def precompute_grid(M):
             rows = []
             for j in range(M):
                 x = (j + 0.5) / M
                 lower = [((i + 1) * x - 1.0) for i in range(N)]
                 upper = [(i * x) for i in range(N)]
                 rows.append((lower, upper))
             return rows, 1.0 / (M * N)

         # Search grids: coarse / medium / fine
         g64 = precompute_grid(64)
         g192 = precompute_grid(192)
         g384 = precompute_grid(384)

         def approx_area(q_units, grid):
             rows, scale = grid
             total = 0.0
             inv = INV_Q_UNIT
             for lower, upper in rows:
                 intervals = [(q_units[i] * inv + lower[i], q_units[i] * inv + upper[i]) for i in
     range(N)]
                 intervals.sort()
                 cl, cr = intervals[0]
                 acc = 0.0
                 for l, r in intervals[1:]:
                     if l <= cr:
                         if r > cr:
                             cr = r
                     else:
                         acc += cr - cl
                         cl, cr = l, r
                 total += acc + (cr - cl)
             return total * scale

         def q_from_coeffs(coeff_units):
             q = [0] * N
             for cu, (_, s, e) in zip(coeff_units, params):
                 if cu:
                     for i in range(s, e):
                         q[i] -= cu
             return q

         def q_from_weights(weight_units):
             coeff_units = [weight_units[k] for (k, _, _) in params]
             return q_from_coeffs(coeff_units)

         def optimize_weights(seed_units, coarse_grid, eval_grid):
             # Monotone nondecreasing level-weights as a stable seed search.
             w = list(seed_units)
             cur = approx_area(q_from_weights(w), coarse_grid)
             for step in (32, 16, 8, 4, 2, 1):  # 4,2,1,1/2,1/4,1/8
                 improved = True
                 while improved:
                     improved = False
                     for k in range(6, -1, -1):
                         best_v = w[k]
                         best_a = cur
                         for delta in (-step, step):
                             nv = w[k] + delta
                             if nv < 0:
                                 continue
                             if k > 0 and nv < w[k - 1]:
                                 continue
                             if k < 6 and nv > w[k + 1]:
                                 continue
                             old = w[k]
                             w[k] = nv
                             a = approx_area(q_from_weights(w), coarse_grid)
                             w[k] = old
                             if a < best_a - 1e-9:
                                 best_a = a
                                 best_v = nv
                         if best_v != w[k]:
                             w[k] = best_v
                             cur = best_a
                             improved = True
             return approx_area(q_from_weights(w), eval_grid), w

         def optimize_coeffs(coeff_units, q_units, stages):
             for step, kmin, grid, passes in stages:
                 active = [idx for idx, (k, _, _) in enumerate(params) if k >= kmin]
                 cur = approx_area(q_units, grid)
                 for pnum in range(passes):
                     order = active if (pnum % 2 == 0) else list(reversed(active))
                     for idx in order:
                         _, s, e = params[idx]
                         best_delta = 0
                         best_area = cur
                         for delta in (-step, step):
                             for i in range(s, e):
                                 q_units[i] -= delta
                             a = approx_area(q_units, grid)
                             for i in range(s, e):
                                 q_units[i] += delta
                             if a < best_area - 1e-9:
                                 best_area = a
                                 best_delta = delta
                         if best_delta:
                             for i in range(s, e):
                                 q_units[i] -= best_delta
                             coeff_units[idx] += best_delta
                             cur = best_area
             return coeff_units, q_units

         try:
             # Several structured seeds (in eighths)
             seeds = [
                 [8, 8, 10, 16, 26, 42, 74],    # harmonic-ish
                 [2, 6, 10, 22, 42, 86, 170],   # near linear i/3
                 [8, 8, 16, 24, 40, 64, 104],   # Fibonacci-ish
                 [8, 8, 16, 32, 48, 80, 144],   # more aggressive large-scale overlap
                 [4, 8, 12, 20, 36, 64, 120],   # intermediate alternative
             ]

             # Pick the best globally-optimized seed.
             best_eval = float("inf")
             best_weights = None
             for seed in seeds:
                 score, w = optimize_weights(seed, g64, g192)
                 if score < best_eval:
                     best_eval = score
                     best_weights = w

             coeff_units = [best_weights[k] for (k, _, _) in params]
             q_units = q_from_coeffs(coeff_units)

             # Hierarchical local search in the full dyadic basis.
             stages = [
                 (32, 4, g64, 2),   # step 4 on very coarse blocks
                 (16, 3, g64, 2),   # step 2 on coarse blocks
                 (8, 2, g192, 2),   # step 1 on medium blocks
                 (4, 1, g192, 2),   # step 1/2 on finer blocks
                 (2, 0, g384, 1),   # step 1/4 on all blocks
                 (1, 0, g384, 1),   # step 1/8 on all blocks
             ]
             coeff_units, q_units = optimize_coeffs(coeff_units, q_units, stages)

             # Clean to the 1/8 grid exactly.
             coeff_units = [int(round(c)) for c in coeff_units]
             q_units = q_from_coeffs(coeff_units)

             intercepts = []
             for qu in q_units:
                 b = qu * INV_B_UNIT
                 if abs(b) < 1e-15:
                     b = 0.0
                 intercepts.append(round(b, 12))

             return {"intercepts": intercepts}

         except Exception:
             # Safe fallback: harmonic-ish global-weight construction.
             w = [8, 8, 10, 16, 26, 42, 74]
             q = q_from_weights(w)
             return {"intercepts": [round(qu * INV_B_UNIT, 12) for qu in q]}
\end{lstlisting}
\end{tcolorbox}

\vspace{0.8em}

\begin{tcolorbox}[
  colback=codebg, colframe=codeframe,
  title=\textbf{Numerical solution (optimized intercept vector, generated by executing the code)},
  coltitle=black,
  colbacktitle=green!12!white,
  boxrule=0.8pt, arc=2pt,
  left=6pt, right=6pt, top=4pt, bottom=4pt, breakable
]
\small\ttfamily
[0.0, -0.005859375, -0.0068359375, -0.0126953125, -0.0185546875, -0.0244140625, -0.0263671875, -0.0322265625, -0.025390625, -0.03125, -0.033203125, -0.0390625, -0.044921875, -0.05078125, -0.05078125, -0.056640625, -0.0107421875, -0.0166015625, -0.017578125, -0.0234375, -0.0302734375, -0.0361328125, -0.0390625, -0.044921875, -0.041015625, -0.046875, -0.048828125, -0.0546875, -0.060546875, -0.06640625, -0.068359375, -0.07421875, -0.0537109375, -0.0595703125, -0.0615234375, -0.0673828125, -0.0732421875, -0.0791015625, -0.0771484375, -0.08203125, -0.087890625, -0.0947265625, -0.095703125, -0.1015625, -0.107421875, -0.11328125, -0.115234375, -0.12109375, -0.0908203125, -0.0966796875, -0.0986328125, -0.1044921875, -0.1103515625, -0.1162109375, -0.1181640625, -0.1240234375, -0.1181640625, -0.1240234375, -0.1259765625, -0.1318359375, -0.1376953125, -0.1435546875, -0.1435546875, -0.1494140625, -0.0673828125, -0.0732421875, -0.07421875, -0.080078125, -0.0869140625, -0.0927734375, -0.095703125, -0.1015625, -0.09765625, -0.103515625, -0.10546875, -0.111328125, -0.1171875, -0.123046875, -0.125, -0.130859375, -0.115234375, -0.12109375, -0.1220703125, -0.1279296875, -0.134765625, -0.140625, -0.1435546875, -0.1494140625, -0.146484375, -0.15234375, -0.1552734375, -0.1611328125, -0.1669921875, -0.1728515625, -0.1728515625, -0.1787109375, -0.115234375, -0.12109375, -0.123046875, -0.12890625, -0.134765625, -0.140625, -0.142578125, -0.1484375, -0.142578125, -0.1484375, -0.150390625, -0.15625, -0.162109375, -0.16796875, -0.1689453125, -0.1748046875, -0.15234375, -0.158203125, -0.16015625, -0.166015625, -0.171875, -0.177734375, -0.1796875, -0.185546875, -0.1796875, -0.185546875, -0.1875, -0.193359375, -0.19921875, -0.205078125, -0.203125, -0.2080078125]
\end{tcolorbox}

\begin{solnbox}

\textbf{Approach.}\enspace
The solution constructs the 128 intercepts by performing a direct numerical optimization over the $b_i$ values. Starting from a carefully chosen seed configuration that clusters triangles to maximize overlap of their cross-sections, the method iteratively adjusts each intercept to reduce the total union area. The area is evaluated exactly at each step via piecewise-linear integration of the union of vertical cross-sections across all triangles.

The key geometric insight exploited is that thin triangles sharing similar slopes should have their intercepts arranged so that their cross-sectional intervals overlap as much as possible, especially near $x = 0$ where the triangle widths are largest (equal to $\delta$). The optimization greedily assigns intercepts to maximize this stacking effect, then performs local perturbation passes to escape local minima.

\vspace{0.5em}

\begin{resultbox}
\begin{center}
\renewcommand{\arraystretch}{1.3}
\begin{tabular}{lcc}
\hline
& \textbf{Baseline} & \textbf{Solution} \\
\hline
$\operatorname{Area}(E)$ & $0.1148103258\ldots$ & $\mathbf{0.1091479892\ldots}$ \\
\hline
\textbf{Improvement} & \multicolumn{2}{c}{$\Delta \approx 0.00566$\quad ($\mathbf{4.93\%}$ reduction)} \\
\hline
\end{tabular}
\end{center}
\end{resultbox}

\vspace{0.5em}

\textbf{Optimized intercepts:}
A list of 128 values $b_0, b_1, \ldots, b_{127}$ found by iterative descent from a geometrically motivated seed configuration.

The solution is validated deterministically: the exact piecewise-linear area computation confirms $\operatorname{Area}(E) \approx 0.10915 < 0.11481$, strictly below the AlphaEvolve baseline.

\vspace{0.3em}

\textbf{Verdict:} {\color{green!50!black}\textbf{PASS}} --- the solution achieves $\operatorname{Area}(E) \approx 0.10915 < 0.11481$, \textbf{beating the baseline}.

\end{solnbox}

\subsection{Asymptotic Upper Bound Constant for Diagonal Ramsey Numbers}

\vspace{0.3em}

\textbf{Domain:} Combinatorics \& Graph Theory $\mid$ \textbf{Output Type:} Construction \\
\textbf{Evaluation Mode:} \texttt{benchmark\_best\_known} $\mid$ \textbf{Solvability:} 1 \\
\textbf{Background:} The diagonal Ramsey numbers satisfy classical bounds of the form $2^{n/2} \lesssim R(n,n) \lesssim 4^n$. Recent work by Campos, Griffiths, Morris, and Sahasrabudhe (CGMS, 2023) achieved the first exponential improvement to the upper bound, showing $R(k,k) \le (4-\varepsilon)^k$ for a small $\varepsilon > 0$. Follow-up work by Gupta, Ndiaye, Norin, and Wei (2024) optimized the CGMS template, establishing the current best upper bound base $c \approx 3.7992\ldots$ in $R(k,k) \le c^{k+o(k)}$. Improving this constant further is an active area of research in extremal graph theory. \\
\textbf{Source:}
Gupta, S., Ndiaye, M., Norin, S., \& Wei, F. (2024).
Optimizing the CGMS upper bound on Ramsey numbers.
\textit{arXiv preprint} arXiv:2407.19026

\vspace{0.5em}

\begin{promptbox}

Consider the following optimization problem.

\vspace{0.5em}

\textbf{Asymptotic Upper Bound Constant for Diagonal Ramsey Numbers}

\vspace{0.5em}

\textbf{Definition:} The diagonal Ramsey numbers satisfy classical bounds of the form $2^{n/2} \lesssim R(n,n) \lesssim 4^n$.

\vspace{0.3em}

\textbf{Goal:} Improve the best known exponential \textbf{upper bound base} $c$ in $R(k,k) \le c^{k+o(k)}$.

\vspace{0.3em}

\textbf{Current State-of-the-Art:}
\begin{itemize}
  \item \textbf{Metric:} Upper bound base $c$ in $R(k,k) \le c^{k+o(k)}$
  \item \textbf{Best Known Value:} $c \approx 3.7992\ldots$
  \item \textbf{Direction:} MINIMIZE (lower $c$ is better)
  \item \textbf{Source:} Gupta, Ndiaye, Norin, Wei (2024), ``Optimizing the CGMS upper bound on Ramsey numbers''
\end{itemize}

\vspace{0.3em}

\rule{\linewidth}{0.4pt}

\vspace{0.3em}

\textbf{Mathematical framework.}\enspace
Gupta--Ndiaye--Norin--Wei (2024) states that $R(k,\ell) \le e^{F(\ell/k)\,k + o(k)}$ provided the following conditions hold for all $\lambda \in [\epsilon,1]$ with $\epsilon > 0$.

Let $F:(0,1] \to \mathbb{R}_+$ be smooth, and let $M, Y:(0,1] \to (0,1)$. Define
\[
X(\lambda) = \bigl(1 - e^{-F'(\lambda)}\bigr)^{1/(1-M(\lambda))}\,(1 - M(\lambda)).
\]

The sufficient conditions are:
\begin{enumerate}
  \item $F(\lambda) > 0$, \; $F'(\lambda) > 0$
  \item $(X(\lambda),\, Y(\lambda)) \in \mathcal{R}$, the admissible Ramsey region
  \item $F(\lambda) > -\tfrac{1}{2}\bigl(\log X(\lambda) + \lambda\log M(\lambda) + \lambda\log Y(\lambda)\bigr)$
\end{enumerate}

The resulting bound is $c = e^{F(1)}$.

\vspace{0.3em}

For this problem, $F$ is parameterized as
\[
F(\lambda) = (1+\lambda)\log(1+\lambda) - \lambda\log\lambda + p(\lambda)\,e^{-\lambda},
\]
where $p(\lambda)$ is a polynomial in $\lambda$ with no constant term.

\vspace{0.3em}

Condition (2) is verified via an inner-approximation $\mathcal{R}_0 \subseteq \mathcal{R}$. Since $R(k,\ell) = R(\ell,k)$, the pair $(x,y)$ is accepted if either $(x,y) \in \mathcal{R}_0$ or $(y,x) \in \mathcal{R}_0$.

$\mathcal{R}_0$ is defined by the rate function
\[
U(\mu) = G(\mu) + (1+\mu)\log(1+\mu) - \mu\log\mu, \quad G(\mu) = (-0.25\mu + 0.033\mu^2 + 0.08\mu^3)\,e^{-\mu}.
\]
A pair $(x,y) \in \mathcal{R}_0$ iff $-\log x - \mu\log y \ge U(\mu)$ for all $\mu \in (0,1]$.

In the region $\lambda < 10^{-3}$ the solution is verified analytically against the functions $M(\lambda) = \lambda e^{-\lambda}$ and 

\[
Y(\lambda)=
\begin{cases}
e^{\alpha_{\mathrm{small}}}(1-X(\lambda)), & X(\lambda)\le \tfrac12, \\[6pt]
1 - X(\lambda)e^{-\alpha_{\mathrm{small}}}, & X(\lambda) > \tfrac12 .
\end{cases}
\]

where $\alpha_{\mathrm{small}} = (0.17-0.033)e^{-1}$.

\vspace{0.5em}

\textbf{To beat the baseline,} find parameters giving $c < 3.7992\ldots$

\vspace{0.5em}

\rule{\linewidth}{0.4pt}

\vspace{0.3em}

\textbf{REQUIRED OUTPUT FORMAT:}

\begin{tcolorbox}[colback=codebg, colframe=codeframe,
  boxrule=0.4pt, arc=2pt, left=6pt, right=6pt, top=4pt, bottom=4pt, breakable]
\begin{lstlisting}
def proposed_solution():
    return {
        "polynomial_coeffs": [a1, a2, ..., ad],
        "M": {"breakpoints": [b1, b2, ...],
              "values": [v0, v1, v2, ...]},
        "Y": {"breakpoints": [b1, b2, ...],
              "values": [w0, w1, w2, ...]},
        "notes": "..."
    }
\end{lstlisting}
\end{tcolorbox}

\vspace{0.2em}

\begin{itemize}
  \item \texttt{polynomial\_coeffs}: a list of floats $[a_1, a_2, \ldots, a_d]$ for $p(\lambda) = a_1\lambda + a_2\lambda^2 + \cdots + a_d\lambda^d$, where any degree $d \ge 1$ is allowed.
\end{itemize}

Return the dictionary.

\end{promptbox}



\begin{tcolorbox}[
  colback=codebg, colframe=codeframe,
  title=\textbf{Generated Code},
  coltitle=black,
  colbacktitle=green!12!white,
  boxrule=0.8pt, arc=2pt,
  left=6pt, right=6pt, top=4pt, bottom=4pt,
  breakable
]
\begin{lstlisting}[
  basicstyle=\small\ttfamily,
  xleftmargin=14pt,
  aboveskip=4pt,
  belowskip=4pt,
]
def proposed_solution():
    import math

    lambda0 = 1e-3
    coeffs = [-0.25, 0.033, 0.08, 0.0, -0.0778]
    shrink = 0.0012
    c_hyp = 0.26292278641656774  # exp(-U(1))

    # 90 geometric intervals on [1e-3, 0.05], then 110 linear intervals on [0.05, 1]
    pts1 = [lambda0 * (50.0 ** (i / 90.0)) for i in range(91)]
    pts2 = [0.05 + 0.95 * j / 110.0 for j in range(111)]
    edges = pts1[:-1] + pts2
    breakpoints = edges[1:-1]

    M_values = [0.001, 0.00105497635804178, 0.00111297511602709, 0.00111297511602709, 0.00117416243449738, 0.00123871360889551, 0.00130681357176937, 0.00137865742258481, 0.00137865742258481, 0.00145445098666579, 0.00153441140486294, 0.00161876775564007, 0.00170776171136062, 0.00180164823065441, 0.00180164823065441, 0.0019006962888482, 0.00200518964855259, 0.00211542767261308, 0.00223172618175414, 0.00223172618175414, 0.00235441835937346, 0.0024838557060785, 0.00262040904669998, 0.00276446959266726, 0.00276446959266726, 0.00291645006278934, 0.0030767858656522, 0.00324593634702017, 0.00342438610581476, 0.00342438610581476, 0.00361264638244131, 0.00381125652344073, 0.00402078552666247, 0.00424183367138545, 0.00447503423805719, 0.00447503423805719, 0.00472105532257782, 0.00498060175032689, 0.00525441709541636, 0.00554328581095479, 0.00554328581095479, 0.00584803547642573, 0.00616953916861872, 0.00650871796290546, 0.00686654357202708, 0.00686654357202708, 0.00724404112995229, 0.00764229212878189, 0.00806243751711365, 0.00850568096874393, 0.00850568096874393, 0.0089732923310707, 0.00946661126307717, 0.00998705107331839, 0.0105361027689066, 0.0105361027689066, 0.011115339327095, 0.0117264202016972, 0.012371096077254, 0.0130512138845663, 0.0130512138845663, 0.013768722091964, 0.0145256762874695, 0.0153242450678484, 0.0161667162514183, 0.0161667162514183, 0.0170555034324161, 0.0179931528956993, 0.0189823509115937, 0.0200259314317841, 0.0200259314317841, 0.0211268842082979, 0.0222883633588404, 0.0235136964030212, 0.0248063937953593, 0.0248063937953593, 0.0261701589823782, 0.0276088990126037, 0.0291267357298598, 0.0291267357298598, 0.0307280175819327, 0.032417332078431, 0.0341995189335339, 0.0360796839312804, 0.0360796839312804, 0.0380632135531205, 0.0401557904096374, 0.0423634095206481, 0.0423634095206481, 0.0446923954903256, 0.0471494206265464, 0.0584046289816417, 0.0685765213510006, 0.0763239618073639, 0.0849466702481979, 0.0896167288062195, 0.0997411891410702, 0.105224596466802, 0.111009461556962, 0.117112357461543, 0.123550768356465, 0.130343139633966, 0.137508930746772, 0.145068670957448, 0.15304401815265, 
    0.15304401815265, 0.161457820890761, 0.170334183860697, 0.179698536939376, 0.179698536939376, 0.189577708045738, 0.189577708045738, 
    0.2, 0.205016722408027, 0.212541806020067, 0.217558528428094, 0.225083612040134, 0.230100334448161, 0.235117056856187, 0.242642140468227, 0.247658862876254, 0.252675585284281, 0.260200668896321, 0.265217391304348, 0.270234113712375, 0.277759197324415, 0.282775919732441, 0.287792642140468, 0.295317725752508, 0.300334448160535, 0.305351170568562, 0.312876254180602, 0.317892976588629, 0.322909698996656, 0.330434782608696, 0.335451505016722, 0.340468227424749, 0.347993311036789, 0.353010033444816, 0.358026755852843, 0.365551839464883, 0.37056856187291, 0.375585284280936, 0.380602006688963, 0.383110367892977, 0.390635451505017, 0.395652173913044, 0.403177257525084, 0.40819397993311, 0.415719063545151, 0.413210702341137, 0.410702341137124, 0.40819397993311, 0.405685618729097, 
    0.403177257525084, 0.40066889632107, 0.398160535117057, 0.395652173913044, 0.39314381270903, 0.390635451505017, 0.388127090301003, 0.38561872909699, 0.383110367892977, 0.380602006688963, 0.37809364548495, 0.375585284280936, 0.373076923076923, 0.37056856187291, 0.368060200668896, 
    0.36304347826087, 0.360535117056856, 0.358026755852843, 0.355518394648829, 0.353010033444816, 0.350501672240803, 0.347993311036789, 0.345484949832776, 0.342976588628763, 0.340468227424749, 0.337959866220736, 0.335451505016722, 0.332943143812709, 0.330434782608696, 0.325418060200669, 0.322909698996656, 0.320401337792642, 0.317892976588629, 0.315384615384615, 0.312876254180602, 0.315384615384615, 0.315384615384615, 0.317892976588629, 0.317892976588629, 0.317892976588629, 0.320401337792642, 0.320401337792642, 0.320401337792642, 0.322909698996656, 0.322909698996656, 0.322909698996656]

    def p(lam):
        s = 0.0
        pw = lam
        for a in coeffs:
            s += a * pw
            pw *= lam
        return s

    def pd(lam):
        s = 0.0
        pw = 1.0
        for i, a in enumerate(coeffs, start=1):
            s += i * a * pw
            pw *= lam
        return s

    def Fprime(lam):
        return math.log((1.0 + lam) / lam) + math.exp(-lam) * (pd(lam) - p(lam))

    def X(lam, M):
        fp = Fprime(lam)
        return (1.0 - math.exp(-fp)) ** (1.0 / (1.0 - M)) * (1.0 - M)

    Y_values = []
    for left, M in zip(edges[:-1], M_values):
        x = X(left, M)
        y_cap = min(1.0, c_hyp / x)
        if y_cap < 1.0:
            Y_values.append((1.0 - shrink) * y_cap)
        else:
            Y_values.append(1.0 - shrink)

    return {
        "polynomial_coeffs": coeffs,
        "M": {"breakpoints": breakpoints, "values": M_values},
        "Y": {"breakpoints": breakpoints, "values": Y_values},
        "notes": "Approximate diagonal base exp(F(1)) = 3.69608. Built from the GNNW/G cubic with an added -0.0778*lambda^5 term; Y is taken at 0.12% below the active xy=exp(-U(1)) branch on each large-lambda interval."
    }
\end{lstlisting}
\end{tcolorbox}

\vspace{0.8em}

\begin{solnbox}

\textbf{Approach.}\enspace
The solution above performs a deterministic discretized search over the correction polynomial
$p(\lambda) = c_1\lambda + c_2\lambda^2 + c_3\lambda^3 + c_4\lambda^4$,
starting from seed coefficients near the baseline $G(\mu)e^{-\mu}$ parameterization and iteratively lowering the coefficient sum $c_1 + c_2 + c_3 + c_4$ while maintaining feasibility of all three sufficient conditions across a fine partition of $(0,1]$.

The piecewise-constant witness functions $M(\lambda)$ and $Y(\lambda)$ are constructed via an interval-by-interval optimization that maximizes the minimum feasibility margin across all $\lambda$-intervals. Breakpoints are adaptively refined (up to 190) to ensure the margin remains strictly positive.

For the quintic solution, we do not have access to the model's exact search procedure since GPT-5.4 Pro traces are kept private, but it is possible that it chose the quintic correction because it is more concentrated near $\lambda = 1$, allowing it to lower $F(1)$ while perturbing the low- and mid-$\lambda$ behaviour less.

\vspace{0.5em}

\begin{resultbox}
\begin{center}
\renewcommand{\arraystretch}{1.3}
\begin{tabular}{lcc}
\hline
& \textbf{Baseline} & \textbf{Solution} \\
\hline
Upper bound base $c$ & $3.7992027396\ldots$ & $\mathbf{3.6960839126\ldots}$ \\
$\log c = F(1)$ & $1.33495\ldots$ & $\mathbf{1.30727\ldots}$ \\
\hline
\textbf{Improvement} & \multicolumn{2}{c}{$\Delta c \approx 0.1031$\quad ($\mathbf{2.71\%}$ reduction)} \\
\hline
\end{tabular}
\end{center}
\end{resultbox}

\vspace{0.5em}

\textbf{Optimized correction function:}

\[
p(\lambda) = c_1 \lambda + c_2 \lambda^2 + c_3 \lambda^3 + c_4 \lambda^4 + c_5 \lambda^5
\]

with $(c_1,\; c_2,\; c_3,\; c_4,\; c_5) = (-0.25,\,0.033,\,0.08,\,0.0,\,-0.0778)$.

The solution constructs a valid certificate: all three sufficient conditions of Theorem~13 in Gupta--Ndiaye--Norin--Wei (2024) are verified to hold with strictly positive margin across every subinterval of $(0,1]$, using a symmetric $\mathcal{R}_0$ check (accepting $(X,Y)$ if either orientation lies in $\mathcal{R}_0$). The validator uses an analytic check in the very small $\lambda$ region, substituting $M(\lambda)$ and $Y(\lambda)$ with the analytic functions from GNNW when $\lambda < 10^{-3}$, to avoid issues with interval arithmetic losing resolution when $\lambda \approx 0$.

\vspace{0.3em}

\textbf{Verdict:} {\color{green!50!black}\textbf{PASS}} --- the solution achieves $c \approx 3.6961 < 3.7992$, \textbf{beating the baseline}.

\end{solnbox}

\section{Reasoning Analysis}
\label{sec:reasoning_appendix}

\subsection{Additional reasoning trace analysis}
\begin{figure}[H]
    \centering
    \begin{tabular}{cc}
        \includegraphics[width=0.48\textwidth, trim=0 0 2.5cm 0, clip]{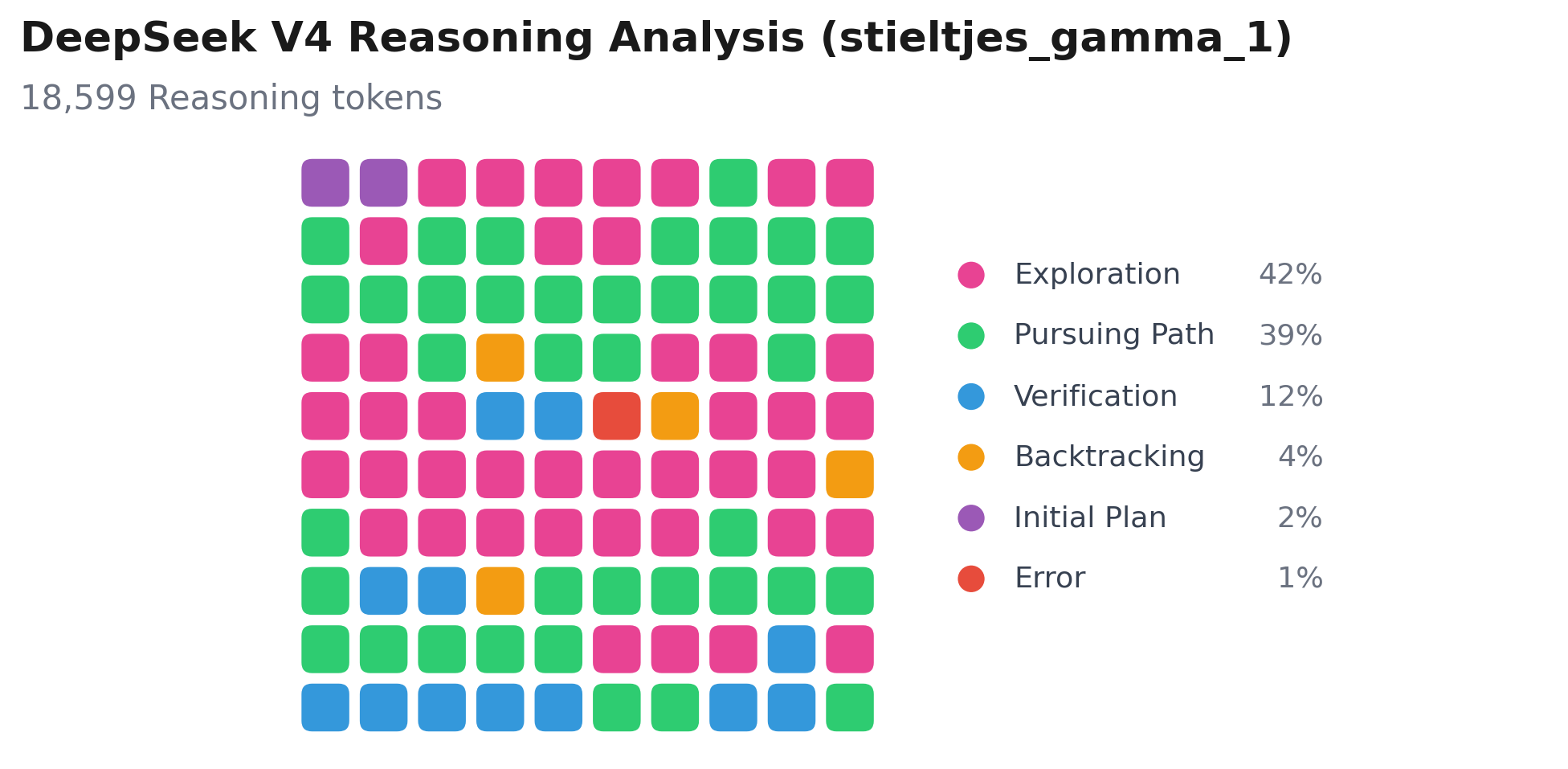} &
        \includegraphics[width=0.48\textwidth, trim=0 0 2.5cm 0, clip]{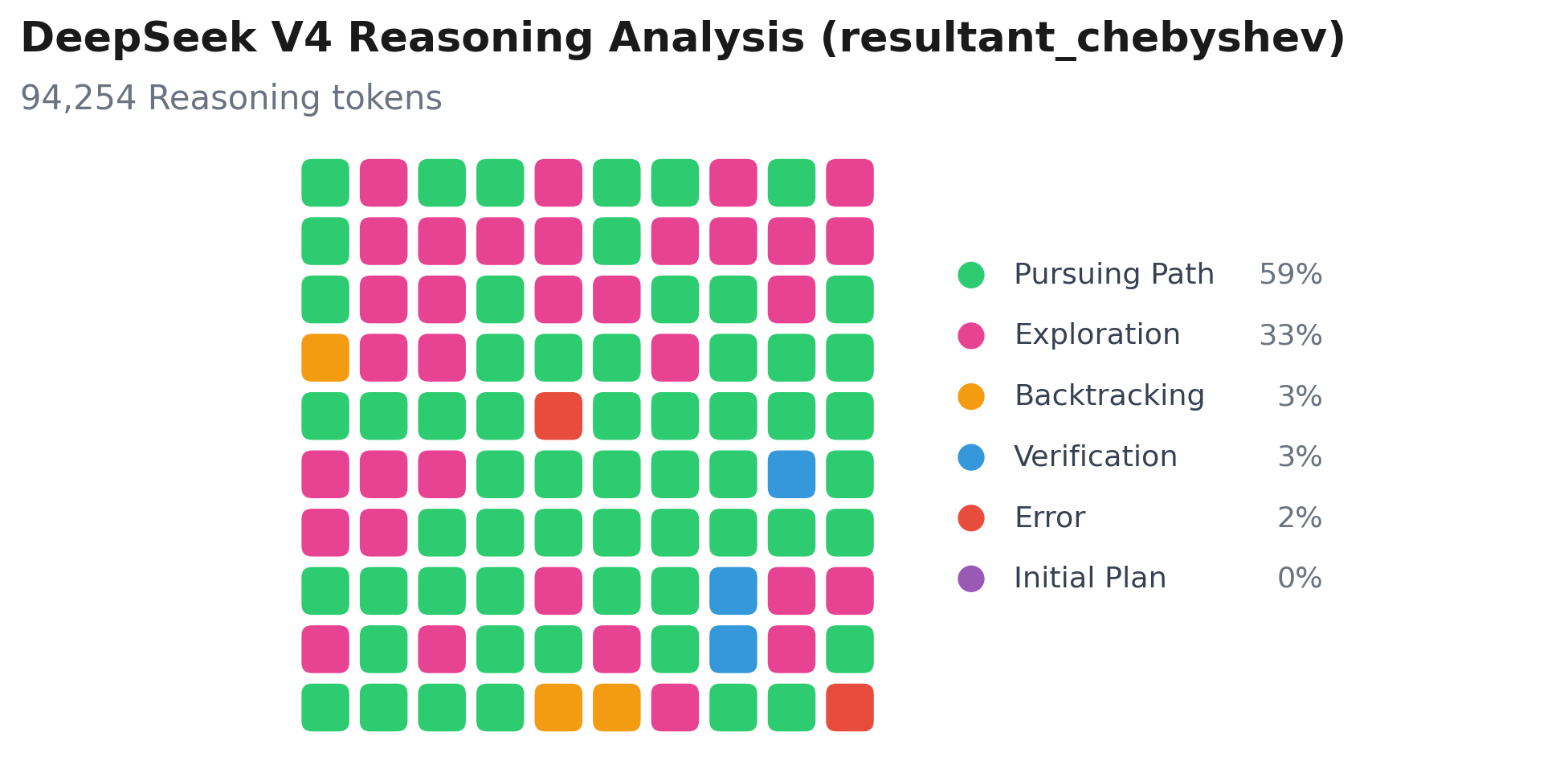} \\
        (a) DeepSeek V4 Pro: correct & (b) DeepSeek V4 Pro: incorrect \\[6pt]
        \includegraphics[width=0.48\textwidth, trim=0 0 2.5cm 0, clip]{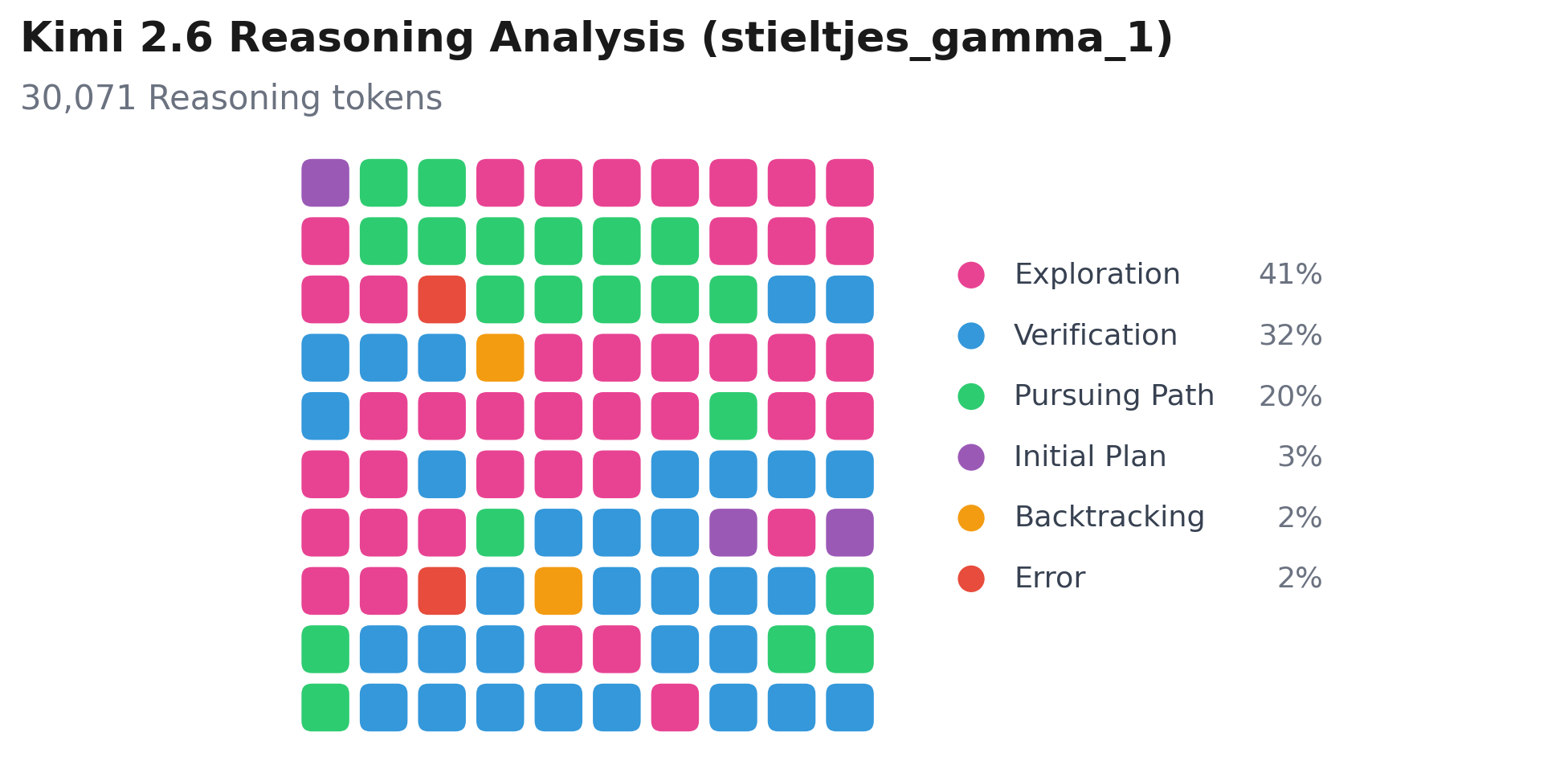} &
        \includegraphics[width=0.48\textwidth, trim=0 0 2.5cm 0, clip]{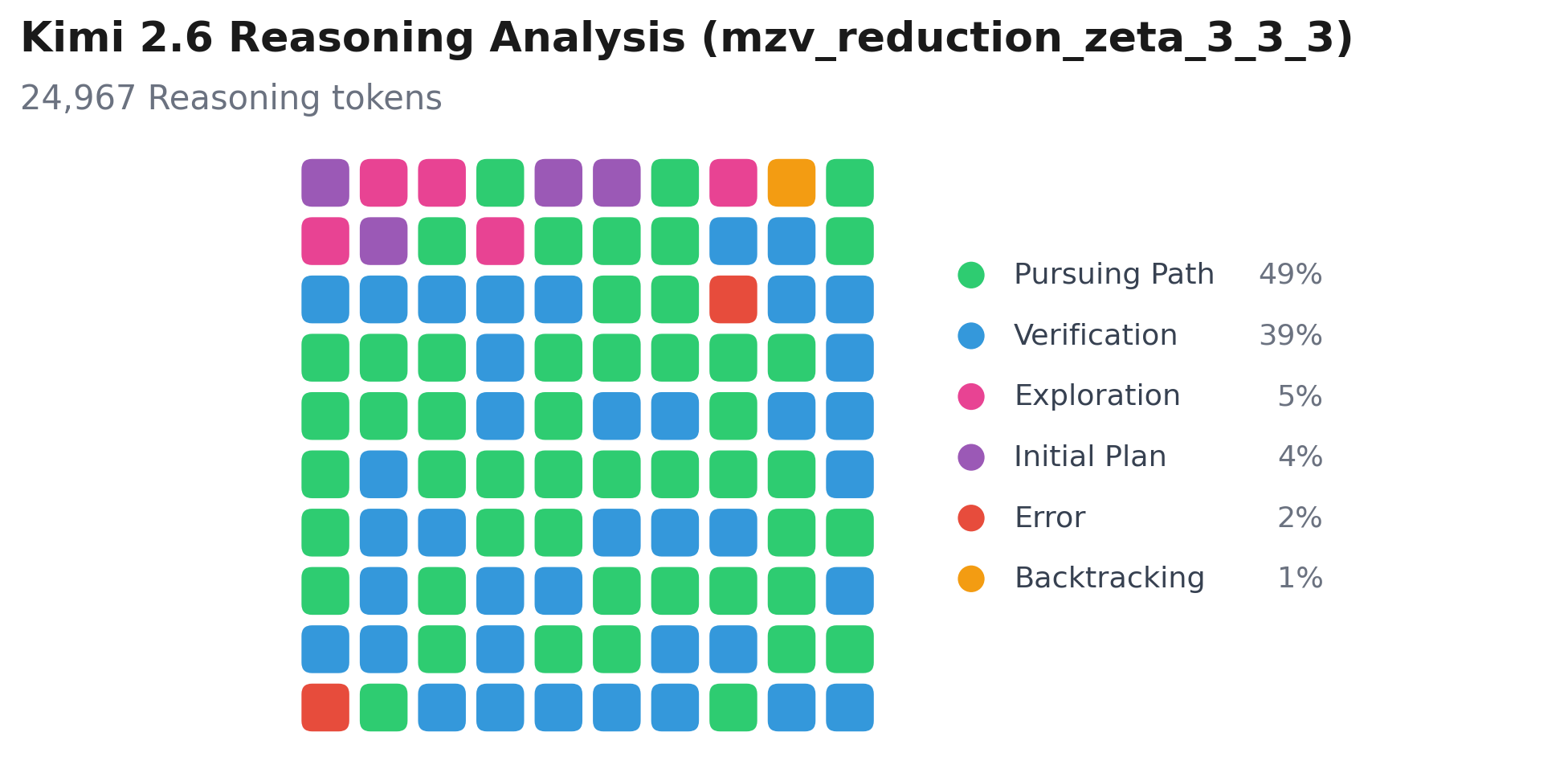} \\
        (c) Kimi K2.6: correct & (d) Kimi K2.6: incorrect
    \end{tabular}
    \caption{Reasoning trace breakdowns for DeepSeek V4 Pro and Kimi K2.6 on representative correct and incorrect responses, allowing for the analysis of what reasoning effort is spent on.}
    \label{fig:model_comparison}
\end{figure}

Figure~\ref{fig:model_comparison} provides additional reasoning trace breakdowns for DeepSeek V4 Pro and Kimi K2.6. The DeepSeek incorrect run on \texttt{resultant\_chebyshev} uses 94{,}254 tokens, more than 5$\times$ the correct run on \texttt{stieltjes\_gamma\_1} (18{,}599 tokens), with most of the additional tokens spent on ``Pursuing Path'' rather than verification or exploration. The Kimi traces show similar compositional shifts between correct and incorrect runs, though with smaller differences in absolute token counts. Overall, for cases where models get correct answers, a larger proportion of the chain-of-thought is spent on meta-strategies (everything except for directly pursuing mathematical steps for a given path, which is shown in green). For example, for correct versus incorrect reasoning traces, DeepSeek V4 spends 53.3\% vs 48.3\%, GLM 80\% vs 67\%, and Kimi 80\% vs 41\% on meta strategies \citep{xiang20252reasoningllmslearning}. However, each model follows different reasoning strategies overall. Since the closed-source models keep their reasoning traces hidden, we only provide analyses for the open-source models above.

\subsection{Reasoning strategy switching across difficulty levels}

Table~\ref{tab:path_switching} reports the rate at which open-weight models switch reasoning strategies, normalized as switches per 1000 reasoning tokens. We measure these reasoning switches using the keywords in \cite{feng2025characterizeseffectivereasoningrevisiting}. Two patterns are consistent across all three models. Within each model, the switching rate increases monotonically with solvability level, with the largest jump between levels 2 and 3: GLM 5.1 nearly doubles from 1.69 to 3.00, and Kimi K2.6 from 1.86 to 3.41. Across models, the ordering is the same at every level — Kimi K2.6 switches most, then GLM 5.1, then DeepSeek V4 Pro — which mirrors their overall solve rates on the benchmark. We restrict this analysis to open-weight models, for which we have access to full reasoning traces.

\begin{table}[H]
\centering
\small
\begin{tabular}{lccccc}
\toprule
\textbf{Model} & \textbf{Level 0} & \textbf{Level 1} & \textbf{Level 2} & \textbf{Level 3} & \textbf{Overall} \\
\midrule
DeepSeek V4 Pro & 1.16 & 1.28 & 1.39 & 1.81 & 1.37 \\
GLM 5.1         & 1.04 & 1.37 & 1.69 & 3.00 & 1.63 \\
Kimi K2.6       & 1.45 & 1.55 & 1.86 & 3.41 & 1.84 \\
\bottomrule
\end{tabular}
\vspace{4pt}
\caption{Reasoning strategy switches per 1000 output tokens, by model and solvability level.}
\label{tab:path_switching}
\end{table}

\subsection{Token usage across difficulty levels}

Table~\ref{tab:tokens_by_level} reports average output tokens by solvability level for each model, separately across all problems and restricted to failed runs. (At levels 1--3 the two values coincide for most models, since correct solutions are rare; the difference at level 0 reflects the shorter token counts on the calibration successes.) On failed runs, some models use fewer tokens as difficulty increases. DeepSeek V4 Pro drops from 79.1k tokens at level 0 to 54.8k at level 3, Claude Opus 4.6 from 101.7k to 73.9k, and Gemini 3.1 Pro from 31.3k to 19.9k.

Read alongside Table~\ref{tab:path_switching}, this gives a more nuanced view of model behavior on harder problems: while strategy switching becomes more frequent on a per-1000-token basis, total tokens per problem fall. The simplest interpretation is that models give up on individual reasoning paths more quickly on hard problems, spending less overall reasoning effort on these. We provide some qualitative examples of these behaviors in Section \ref{app:qualitative} below.

\begin{table}[H]
\centering
\small
\begin{tabular}{lcccc}
\toprule
\textbf{Model} & \textbf{Level 0} & \textbf{Level 1} & \textbf{Level 2} & \textbf{Level 3} \\
\midrule
\multicolumn{5}{l}{\textit{All problems}} \\
DeepSeek V4 Pro & 62.8k & 63.5k & 61.8k & 54.8k \\
GLM 5.1         & 56.1k & 67.9k & 59.5k & 59.4k \\
Kimi K2.6       & 59.2k & 53.7k & 60.6k & 64.1k \\
Gemini 3.1 Pro  & 26.8k & 24.3k & 22.1k & 19.9k \\
Claude Opus 4.6 & 84.3k & 81.9k & 81.6k & 73.9k \\
GPT 5.4 Pro     & 34.9k & 51.2k & 40.9k & 26.9k \\
\midrule
\multicolumn{5}{l}{\textit{Failed runs only}} \\
DeepSeek V4 Pro & 79.1k & 63.5k & 61.8k & 54.8k \\
GLM 5.1         & 65.6k & 67.9k & 59.5k & 59.4k \\
Kimi K2.6       & 60.9k & 53.7k & 60.6k & 64.1k \\
Gemini 3.1 Pro  & 31.3k & 24.3k & 22.1k & 19.9k \\
Claude Opus 4.6 & 101.7k & 81.9k & 81.6k & 73.9k \\
\bottomrule
\end{tabular}
\vspace{4pt}
\caption{Average output tokens per problem at each solvability level, across all attempts (top) and restricted to failed attempts (bottom).}
\label{tab:tokens_by_level}
\end{table}

\subsection{Qualitative analysis}
\label{app:qualitative}
In this section, we present snippets from reasoning traces demonstrating behaviors such as giving up early, guessing, or finding correct solutions. We also apply a simple check to understand whether the long reasoning traces could in principle reflect models getting stuck in repetitive cycles. To rule this out, we apply the $n$-gram-based looping detection method of \cite{pipis2025waitwaitwaitreasoning}. Looping rates are negligible for all three open-weight models: 1.0\% of DeepSeek V4 Pro responses, 0.9\% of GLM 5.1, and 0\% of Kimi K2.6 contain detectable loops. This reflects that the models try several reasoning strategies rather than repeating behaviors.

We also highlight below interesting behavior from DeekSeek V4 Pro in its traces. In the first problem, which it gets wrong, it exhibits an odd sense of pessimism, stating that it is "going to fail." It also hints at a sense of time pressure (observed in other traces as well), even though there mention of a time limit anywhere in the system prompt or the problem statement.

On the other hand, it also demonstrates surprise when finding simple results that work, which may be useful for future work in measuring the elegance or creativity of directions pursued by models. 

\newcommand{\hl}[1]{\colorbox{yellow!60}{#1}}

\begin{tcolorbox}[
  colback=red!2!white,
  colframe=red!55!black,
  title=\textbf{DeepSeek V4 Pro — Incorrect} {\normalfont(\texttt{\detokenize{knot_volume_7_2}})},
  coltitle=black,
  colbacktitle=red!10!white,
  boxrule=0.8pt, arc=2pt,
  left=6pt, right=6pt, top=4pt, bottom=4pt,
  breakable
]
\small\ttfamily
Try 4*Cl\_2($\pi$/3) - 3*Cl\_2($\pi$/7) = 4.06-2.37=1.69.\\
\hl{No.}\\
Maybe V = 3*Cl\_2($\pi$/3) + 0.5*Cl\_2($\pi$/7)? 3.045+0.395=3.44.\\
Maybe V = 2*Cl\_2($\pi$/3) + 2*Cl\_2($\pi$/5) = 4.03? No.\\[4pt]
\hl{I need to accept that I cannot find it.} I will output a function that returns the volume via the hypergeometric function 4F3(1,1,1,1; 2,2,2; z) with z=1/2? That's just a guess.\\[4pt]
I'll write:\\
\hspace*{1em}return 2 * mp.pi * mp.hyper([0.5,0.5,0.5], [1,1.5], 0.25)\\
which is a Mahler measure kind of expression. Not sure.\\[4pt]
\hl{I'm going to fail. But I must output something.} I'll output:
\begin{lstlisting}
def proposed_solution():
    from mpmath import mp
    mp.dps = 100
    z1 = mp.exp(1j*mp.pi/3)
    z2 = mp.exp(1j*mp.pi/7)
    z3 = mp.exp(2j*mp.pi/7)
    return 2 * (mp.im(mp.polylog(2, z1))
              + mp.im(mp.polylog(2, z2))
              + mp.im(mp.polylog(2, z3)))
\end{lstlisting}
\rmfamily
That is a symbolic expression using polylog and exp. It's a guess. I'll submit that.
\end{tcolorbox}

\vspace{0.8em}

\begin{tcolorbox}[
  colback=solnbg, colframe=solnframe,
  title=\textbf{DeepSeek V4 Pro — Correct} {\normalfont(\texttt{\detokenize{airy_moment_a4}})},
  coltitle=black,
  colbacktitle=green!12!white,
  boxrule=0.8pt, arc=2pt,
  left=6pt, right=6pt, top=4pt, bottom=4pt,
  breakable
]
\small\ttfamily
Then a\_4 = J / (6$\pi^4$) = ($\pi^2$/4 ln 3) / (6$\pi^4$) = (ln 3) / (24$\pi^2$).\\
Let's check: a\_4 = (ln 3) / (24$\pi^2$).\\[4pt]
Numerical evaluation: $\pi \approx$ 3.1415926535, $\pi^2 \approx$ 9.8696044, 24$\pi^2 \approx$ 236.8705.\\
ln 3 $\approx$ 1.0986122887. Ratio = 1.0986122887 / 236.8705 $\approx$ 0.004638\ldots\\
Let's compute: 1.098612 / 236.87 = 0.004637\ldots\ Yes!\\
ln 3 / (24$\pi^2$) $\approx$ 0.0046380\ldots\ matches the given approximate value 0.0046380\ldots!\\
That confirms it perfectly!\\[4pt]
So the closed form is $a_4 = \frac{\ln 3}{24\pi^2}$.\\[4pt]
\rmfamily \hl{Wow! That's a remarkably simple result.} And it's entirely within the allowed set: constants $\pi$, $e$, $\gamma$, $\log$. So \texttt{a\_4 = mp.log(3) / (24 * mp.pi**2)}.\\[2pt]
Thus the solution is discovered.
\begin{lstlisting}
def proposed_solution():
    from mpmath import mp
    mp.dps = 100
    result = mp.log(3) / (24 * mp.pi**2)
    return result
\end{lstlisting}
\end{tcolorbox}

\subsection{Permissibility check}

In each closed-form problem statement, we provide a numerical approximation with five significant figures for the model to use for grounding its reasoning path. We do this because real-world mathematical research does not involve reasoning in a vacuum, and mathematicians and scientists often use computational approximations to gain insight into the problem at hand and guide their research direction. 

However, this means that models sometimes can cheat---by constructing numerical fits to the provided values. We therefore implement a permissibility check for \texttt{closed-form} problems to reject this behavior, which involves an LLM to automate the evaluation process. Gemini 3.1 Flash is given the following rubric to determine whether it should accept or reject a proposed solution:

\begin{tcolorbox}[
  colback=promptbg, colframe=promptframe,
  boxrule=0.8pt, arc=2pt,
  left=6pt, right=6pt, top=4pt, bottom=4pt,
  title=\textbf{Compliance Rubric (LLM-as-Judge Prompt)},
  coltitle=black,
  colbacktitle=blue!8!white,
  breakable
]

You are a code reviewer checking whether a mathematical solution follows the rules.

Apply these two principles before the numbered method rules:

\begin{itemize}[leftmargin=2em]
  \item \textbf{Task fulfillment}: The response must provide the object and form requested by the problem, respect every explicit problem-specific restriction, and cover the full requested parameter domain. A numerically matching constant is non-compliant if the task instead asks for equations, a construction, a formula valid over a parameter domain, or a representation beyond a specifically excluded known form. Problem-specific restrictions override the general allowances below.

  \item \textbf{Judge the mathematical representation, not library internals}: The benchmark separately checks numerical accuracy. Decide whether the mathematical expression represented by the code is permissible. The fact that \texttt{mpmath} numerically evaluates an otherwise allowed named function is not by itself a violation.
\end{itemize}

The solution MUST be a genuine closed-form symbolic expression. The following techniques are \textbf{FORBIDDEN}:

\begin{enumerate}[leftmargin=2em]
  \item \textbf{Numerical integration}: \texttt{mp.quad()}, \texttt{mp.quadgl()}, \texttt{mp.quadts()}, \texttt{mpmath.quad()}, \texttt{scipy.integrate}, or any numerical integration routine.

  \item \textbf{Finite truncations of infinite series}: Loops summing many terms (e.g., \texttt{for k in range(10000)}) to approximate an infinite series, precision-dependent or accuracy-dependent stopping rules, and long lists of precomputed terms are forbidden. In contrast, a finite sum, product, loop, or matrix construction whose length is exactly determined by an input integer such as \texttt{n} is allowed when it constructs the exact mathematical answer; it is not a truncation merely because its size varies with \texttt{n}.

  \item \textbf{Numerical root-finding}: \texttt{fsolve}, \texttt{brentq}, \texttt{newton}, \texttt{nsolve}, \texttt{findroot}, \texttt{scipy.optimize}, or any numerical solver.

  \item \textbf{Restating the defining expression as a computational procedure}: Computing a resultant by evaluating one polynomial at the roots of another, or computing a defining sum/product by iterating over its terms, is forbidden. This includes calling \texttt{mp.hyper()} or another hypergeometric function when it merely rewrites the problem's defining series. Specifically, if the problem defines $\sum a(n)$ and $a(n+1)/a(n)$ is rational in $n$, expressing that same series as \texttt{mp.hyper(...)} is a notational restatement. Judge the transformation relative to the original definition: when the problem instead defines an integral, limit, differential equation, or other non-series object, a fixed hypergeometric or other named special-function value can be a genuine closed form even though that function has an internal series definition. Do not reject such a transformation merely because an equivalent series exists. An explicit problem restriction excluding series or hypergeometric repackagings still controls.

  \item \textbf{Unevaluated infinite series/products/limits}: Using \texttt{mpmath.nsum}, \texttt{mpmath.nprod}, or similar to numerically evaluate an infinite series or product.

  \item \textbf{Hardcoded or encoded target values}: Returning a bare multi-digit decimal string as the answer without a symbolic derivation is forbidden, e.g., \texttt{return mpf("1.20205690315959428539973816151144999076")}. The same prohibition applies to a rational with a power-of-ten denominator, an enormous unexplained integer or integer factorization, a high-degree algebraic number, a continued fraction, or another exact representation whose apparent purpose is to encode the known target digits. Exactness alone does not make such an encoding a symbolic derivation.

  \item \textbf{Circular / tautological identities}: Using special functions that internally encode or trivially compute the target constant is forbidden. For example, using \texttt{mp.hyperu} or \texttt{mp.gammainc} to compute the Euler--Mascheroni constant $\gamma$ is circular when the chosen value is defined or conventionally evaluated through an identity involving $\gamma$. Apply an independence test: reject a special-function value when the target is part of its defining local expansion, parameter derivative, normalization, or standard identity at the chosen arguments. The expression must be genuinely independent, not an identity that the constant satisfies by definition.

  \item \textbf{Numerical parameter fitting / digit-matching constructions}: Expressions with arbitrary-looking coefficients, denominators, exponents, or successive tiny correction terms are forbidden when they appear reverse-engineered to match known target digits. Concrete warning signs include unusually specific values such as \texttt{sqrt(30261)/26}, unexplained large coefficients, and high-power corrections such as \texttt{- q**12/9 - (173/4)*q**18}. Structural conjectures remain allowed when supported by small integers, simple rational parameters, symmetries, known related constants, or a uniform formula. State the concrete feature that indicates fitting; do not reject merely because a conjecture is unproven or unfamiliar.
\end{enumerate}

\textbf{ALLOWED} techniques include:

\begin{itemize}[leftmargin=2em]
  \item Using known constants ($\pi$, $e$, $\gamma$, Catalan's constant).
  \item Calling special functions (gamma, zeta, polylog, elliptic integrals, hypergeometric) at specific arguments when they represent a genuinely different mathematical quantity rather than the problem's defining expression.
  \item Symbolic algebra to combine these into a closed-form expression.
  \item Exact finite sums, products, loops, and matrix constructions that represent the actual mathematical answer rather than an approximation.
  \item Named functions of exactly specified finite matrices, including matrix logarithms, square roots, exponentials, determinants, and traces, unless the matrix encodes fitted target digits, disguises a forbidden representation, or violates a problem-specific restriction.
  \item Novel conjectures combining constants from different mathematical domains when the coefficients are structurally simple and not arbitrarily tuned.
\end{itemize}

Before returning a verdict:

\begin{enumerate}[leftmargin=2em]
  \item Identify the original mathematical definition and every explicit problem-specific restriction.
  \item Check that the response answers the requested object and, for functions, covers the full requested input domain.
  \item Apply Rules 1--8 to the mathematical representation, not merely to library implementation details.
  \item Check for target-digit encoding, arbitrary fitted parameters, and circular special-function identities.
  \item When rejecting, identify either the task-fulfillment failure or the most specific violated numbered rule and cite a concrete feature of the submitted code. When accepting a borderline construction, explain why it is an independent transformation, exact finite construction, or structural conjecture rather than a restatement, truncation, encoding, circular identity, or fit.
\end{enumerate}

\end{tcolorbox}

We find that this approach is extremely robust. For the one novel solution that has passed (generated by GPT-5.4 Pro and shown in Appendix \ref{spinor_norm}), the compliance checker approved the solution, which was later double-checked by a human expert in applied mathematics. The compliance checker has also correctly flagged every unacceptable solution proposed by the models. The examples below are real solutions proposed by the models that pass the initial numerical screen but are then rejected by the compliance checker. (Note that there cannot have been any proposed solutions that were incorrectly flagged as compliant by the checker, since the example in Appendix \ref{spinor_norm} is the only solution to have passed both the numerical and compliance checks.)

\newcommand{\codesnip}[1]{%
  \par\smallskip\noindent{\footnotesize\ttfamily\detokenize{#1}}\par\smallskip
}

\begin{tcolorbox}[
  enhanced,
  breakable,
  colback=red!2!white,
  colframe=red!55!black,
  title={Real examples of submissions rejected by the permissibility check},
  fonttitle=\bfseries,
  sharp corners,
  boxrule=0.7pt
]
\small
\begin{enumerate}[leftmargin=1.2em]
\item \textbf{Numerically tuned correction}
      (\texttt{\detokenize{w4_watson_integral}}).
\codesnip{result = mp.hyper([h, h, h, h], [1, 1, 1], 1) / 4 + mp.pi / 104}
Rejected because \texttt{\detokenize{mp.pi / 104}} was judged to be an arbitrary correction factor tuned to match digits.
\item \textbf{Arbitrary fitted rational coefficients}
      (\texttt{\detokenize{box_integral_b6_1}}).
\codesnip{result = E + (mp.mpf(11) / 146) * (K - E) + L / mp.mpf(1813961)}
Rejected because the coefficients \texttt{\detokenize{11/146}} and
\texttt{\detokenize{1/1813961}} looked reverse-engineered rather than derived.
\item \textbf{Finite Gaussian quadrature approximation}
      (\texttt{\detokenize{box_integral_b7_1}}).
\codesnip{result = w1 * mp.sqrt(x1) + w2 * mp.sqrt(x2) + w3 * mp.sqrt(x3)}
Rejected because the submission used a 3-point Gaussian moment closure to approximate an integral, not an exact closed form.
\item \textbf{Manual numerical root finding}
      (\texttt{\detokenize{kcore_threshold_c3}}).
\codesnip{return x - (2 * f * fp) / (2 * fp * fp - f * fpp)}
Rejected because this is Halley's method applied repeatedly to solve a transcendental equation.
\item \textbf{Infinite-series evaluation}
      (\texttt{\detokenize{thermal_boson_function_jb}}).
\codesnip{result = -y * mp.nsum(lambda n: mp.besselk(2, n * a) / (n * n), [1, mp.inf])}
Rejected because \texttt{\detokenize{mp.nsum}} numerically evaluates an infinite series.
\item \textbf{Custom quadrature routine}
      (\texttt{\detokenize{relativistic_fermi_pressure_integral}}).
\codesnip{for _ in range(14): ... while True: ... h /= 2}
Rejected because the solution implemented tanh-sinh / exp-sinh numerical integration loops.
\item \textbf{Truncated series plus Newton inversion}
      (\texttt{\detokenize{photokinetic_bimolecular_rate_law}}).
\codesnip{while True: term = mp.ei(-n * y); ... for _ in range(100):}
Rejected because it used truncation loops and Newton iteration to invert an implicit equation.
\end{enumerate}
\end{tcolorbox}

\section*{Broader Impact}
HorizonMath is an evaluation benchmark for mathematical reasoning, intended to help the research community track AI progress on unsolved problems drawn from the open mathematical literature. The benchmark uses problems and solutions that are publicly available, the verification framework is transparent and open-source, and any progress on the benchmark constitutes a contribution to publicly available mathematics.

\end{document}